\documentclass[11pt]{article}

\usepackage{macros}
\usepackage{comment}
\usepackage{multirow}

\DeclareRobustCommand{\onePLOTnat}{\textsc{1-stage PLOT native}}
\DeclareRobustCommand{\onePLOTpca}{\textsc{1-stage PLOT PCA}}
\DeclareRobustCommand{\twoPLOTnat}{\textsc{2-stage PLOT native}}
\DeclareRobustCommand{\twoPLOTpca}{\textsc{2-stage PLOT PCA}}

\title{\textsc{PLOT}: Progressive Localization via Optimal Transport\\
in Neural Causal Abstraction}
\author{
Jonathn Chang \quad Arya Datla \quad Ziv Goldfeld\\[1.1em]
Cornell University
}
\date{}

\begin{document}

\blfootnote{Emails: \texttt{\{jc3683, avd42, goldfeld\}@cornell.edu}. Code is available at \url{https://github.com/jchang153/causal-abstractions-ot}.}
\maketitle

\begin{abstract}
Causal abstraction offers a principled framework for mechanistic interpretability, aligning a high-level causal model with low-level neural computation through interchange intervention analysis. Finding such an alignment, however, often requires fitting and evaluating separate learned mappings across many candidate neural locations. We introduce PLOT, a gradient-free approach for joint correspondence discovery via a global matching of intervention effects. PLOT represents abstract and neural interventions by geometric signatures of their effects on the shared task output and fits an optimal transport coupling between the two collections. The coupling can be calibrated directly into an executable intervention handle or used progressively to select coarse parent sites for finer localization, reducing the cost of searching broad collections of neural sites. Experiments on hierarchical equality, binary addition, and multiple-choice question answering (MCQA) demonstrate fast and accurate direct handles without learning intervention rotations and show that progressive localization can improve accuracy under intervention-size constraints while reducing computational cost. PLOT can also guide gradient-based subspace learning, with PLOT-guided distributed alignment search (DAS) attaining accuracy comparable to full DAS at approximately $3.6\times$ and $20\times$ lower serial runtime on binary addition and MCQA, respectively. PLOT thus separates the search for candidate correspondences from optional subspace learning and provides a common framework for constructing and testing neural intervention handles for a given causal model.

\end{abstract}

\section{Introduction}

Causal abstractions of neural models are central to mechanistic interpretability. Given a trained neural network and a high-level model with interpretable latent variables, the goal is to determine whether the network realizes the abstract computation \citep{pearl2009causality,geiger2025causal}. Under this view, interpretation is not merely finding correlated neurons or salient features. Rather, an abstract variable is realized only if interventions on it can be matched by interventions at internal neural sites. This turns interpretability into a causal localization problem of finding a faithful correspondence between abstract variables and their high-dimensional, distributed neural realization.

Recent work has made this question operational through interchange interventions and distributed alignment search (DAS) \citep{geiger2022iit,geiger2024das,wu2024boundlessdas}. DAS fixes a neural site, such as a layer, and learns a rotation under which swapping a selected subspace between source and base activations reproduces the abstract intervention. This makes DAS expressive and accurate \citep{mueller2025mib}, but finding the relevant location can require fitting and evaluating alignments across many neural sites. Thus, DAS is powerful once a site has been proposed, while correspondence discovery across a broad candidate family remains a separate problem. We address this problem via a joint matching between the intervention effects of abstract variables and candidate neural sites.

\subsection{Contributions}\label{subsec:contributions}

We propose PLOT, a gradient-free approach for joint correspondence discovery via a global matching of intervention effects. Abstract and neural states may live in different spaces, but interventions on them can be compared through their effects on the common task output. PLOT records these effects across base/source pairs as geometric signatures and fits an optimal transport (OT) coupling between the abstract and neural signature collections. The coupling gives a global soft correspondence between high-level variables and candidate neural sites that can be calibrated into executable handles, restrict the candidate family for subsequent localization, or guide gradient-based subspace learning.

\begin{figure}[t!]
    \centering
    \vspace{-2mm}
    \includegraphics[width=.85\linewidth]{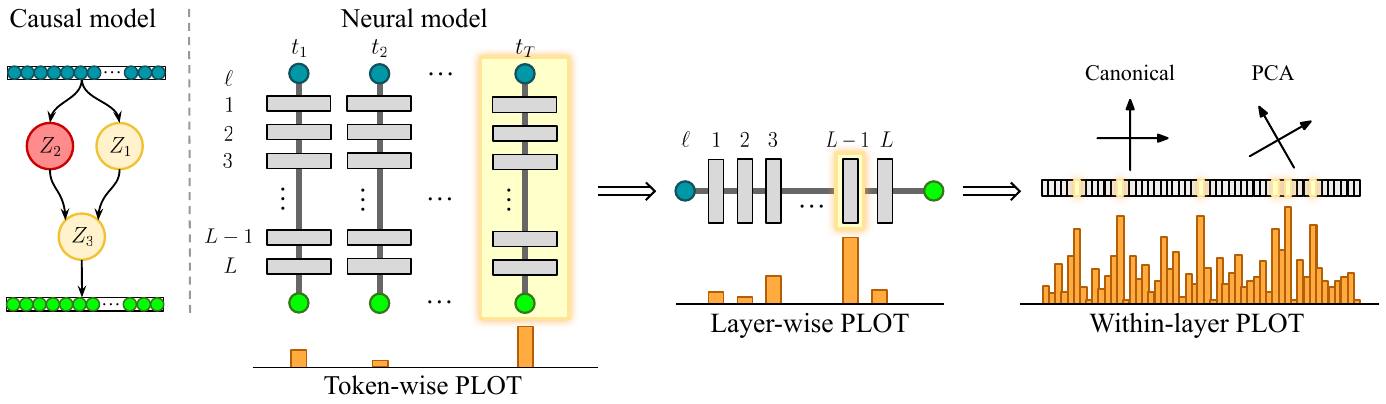}
    \vspace{-2mm}
    \caption{Progressive localization via optimal transport. The diagram follows one high-level variable, $Z_2$ in red, although each OT stage jointly matches all target variables to the neural site family. PLOT first localizes coarse sites, such as tokens/layers, then refines within them in native or PCA~coordinates.} 
    \label{fig:plot-hierarchy}
    \label{fig:PLOT_diagram}
    \vspace{-4mm}
\end{figure}

The neural candidate family can range from individual neurons or coordinate blocks to PCA spans or entire activation vectors, and may include sites from multiple layers or timesteps. PLOT can match all such candidates jointly, allowing the calibrated handle to combine sites across neural locations. When the candidate family contains many irrelevant sites, we use one-sided unbalanced OT (UOT), which fixes the abstract marginal while relaxing the neural marginal so that irrelevant candidates need not absorb transport mass. PLOT can also exploit a natural hierarchy among the candidate sites through progressive localization. A coarse coupling first identifies promising parent sites, after which the matching is refined only within the selected locations. As illustrated in \Cref{fig:plot-hierarchy}, this can proceed from coarse sites such as tokens or layers to within-layer coordinate groups or PCA spans. The progressive procedure can improve the accuracy of compact handles while reducing localization cost or can be used to guide local subspace methods such as DAS.

Empirically, we study PLOT on three tasks of increasing localization complexity. On the hierarchical equality benchmark (HEQ) of \citet{geiger2024das}, single-stage PLOT over MLP coordinates recovers accurate, compact handles without learned rotations or gradient-based optimization. On 4-bit binary addition with a GRUCell backbone, one- and two-stage PLOT recover accurate carry handles, with PCA outperforming native coordinates and two-stage localization improving accuracy and runtime. PLOT-guided DAS matches full-DAS accuracy at roughly $3.6\times$ lower runtime. Finally, on the multiple-choice question-answering (MCQA) benchmark \citep{mueller2025mib} with Gemma-2-2B \citep{gemma2}, we compare direct PLOT handles against masking- and patching-based baselines, as well as PLOT-guided DAS and full DAS. Two-stage PLOT improves over one-stage matching under matched handle-dimension limits while substantially reducing runtime, and its gradient-free PCA handles are within only 3.5 percentage points of full DAS accuracy. PLOT-guided DAS remains close to full DAS while reducing runtime by roughly $20\times$.


\subsection{Literature Review}\label{subsec:lit_review}

A broad line of mechanistic interpretability work seeks to localize internal structures that mediate specific model behaviors, spanning factual associations \citep{meng2022rome}, learned features \citep{cunningham2023sae}, and algorithmic circuits \citep{mueller2025mib,polyakov2025interpretability}. These methods are often geared toward feature discovery, model editing, or benchmark-specific analysis rather than a general causal correspondence criterion. The causal abstraction framework sharpens this target by asking whether an intervention on a hypothesized high-level variable can be reproduced by an internal intervention in the network \citep{geiger2025causal}. Methods developed for this framework include interchange intervention training (IIT) \citep{geiger2022iit}, DAS \citep{geiger2024das}, and subsequent methods that automate parts of the DAS search \citep{wu2024boundlessdas,sun2025hyperdas}. PLOT instead constructs a joint correspondence by matching the output effect signatures of abstract variables and candidate neural sites. The resulting coupling can directly define intervention handles or be applied progressively to restrict subsequent localization and guide local subspace methods such as DAS. Related approaches based on PCA \citep{tigges2023,pca_ref}, sparse autoencoders (SAEs) \citep{bricken2023monosemanticity,huben2024sparse}, and masking \citep{chaudhary2024dbm,mueller2025mib} provide candidate representations and additional localization baselines.

OT has also been used to align causal models across levels of abstraction. Most notably, causal OT of abstractions (COTA) \citep{felekis2023cota} formulates abstraction between causal models at different granularities as a multimarginal OT problem \citep{gangbo1998optimal}. COTA enforces causal consistency through do-calculus constraints and intervention-aware transport costs. This setting, which aligns two explicit causal models, differs from PLOT, which uses OT to align a high-level causal model with the internal computation of a neural network. More broadly, our work is also related to causal OT and OT-based distances for causal models \citep{acciaio2016causal,lassalle2018causal,cheridito2023optimal}, though these works do not study neural causal abstraction.

\section{Background and Preliminaries}
\subsection{Causal Abstraction of Neural Computation}\label{subsec:background_causal}

Causal abstraction in mechanistic interpretability \citep{pearl2009causality,geiger2025causal} compares an abstract causal model $\mathfrak{C}:\mathcal{X}\to\mathcal{Y}$, with interpretable variables $Z_1,\ldots,Z_m$, to a trained neural model $\mathfrak{N}:\mathcal{X}\to\mathcal{Y}$ that solves the same task. Formal notions such as exact transformations require consistency under corresponding interventions after mapping low-level states to high-level states \citep{rubenstein2017causalconsistencystructuralequation,beckers2019abstractingcausalmodels}. Constructive abstractions additionally require this map to factor over disjoint groups of low-level variables. Here, we consider the neural localization setting of \citet{geiger2021causalabstractions,mueller2025mib}, in which proposed correspondences yield intervention handles that are assessed through \emph{interchange interventions}.

\paragraph{Abstract swap.}
Given a base/source pair of inputs $(x^b,x^s)\in\mathcal{X}^2$ and an abstract
variable $Z_i$, the corresponding abstract counterfactual output is $y^{\mathrm{abs\_swap}}_i(x^b,x^s)\coloneqq \mathfrak{C}\left(x^b;\operatorname{do}\bigl(Z_i\leftarrow Z_i(x^s)\bigr)\right)$. This intervention keeps the base input fixed while replacing only the value of $Z_i$ by its source value and recomputing the downstream variables. The operation is an interchange intervention, while its result is a base-specific counterfactual outcome.

\paragraph{Neural swap.}
A neural site $s_j$ specifies a location and a collection of coordinates or features to intervene on, with values $s_j(x)$ on input $x$. A site may comprise a full layer activation vector, selected PCA directions or SAE features, a neuron subset selected by DBM \citep{chaudhary2024dbm}, or a $k$-dimensional rotated subspace learned by DAS \citep{geiger2024das}. The analogous neural interchange output is $y^{\mathrm{nn\_swap}}_j(x^b,x^s)\coloneqq\mathfrak{N}\left(x^b;\operatorname{do}\bigl(s_j\leftarrow s_j(x^s)\bigr)\right)$. 
For coordinate and orthogonal subspace swaps, replacement leaves the complementary coordinates at that layer unchanged.


\paragraph{Correspondence criterion.}
For each abstract variable $Z_i$, an intervention handle $H_i$ specifies which neural sites to modify and how. Writing $y_{H_i}^{\mathrm{nn\_swap}}(x^b,x^s)$ for its output, \emph{interchange intervention accuracy} (IIA) is the fraction of evaluated base/source pairs satisfying $\smash{y_{H_i}^{\mathrm{nn\_swap}}(x^b,x^s)\sim y_i^{\mathrm{abs\_swap}}(x^b,x^s)}$, where $\sim$ denotes task-dependent output agreement \citep{geiger2024das,mueller2025mib}. A handle specifies an intervention rather than a map from neural state values to abstract values. High empirical IIA supports the proposed correspondence on the evaluated pairs, but does not by itself establish an exact or constructive abstraction.

\subsection{Optimal Transport}

\begin{wrapfigure}{r}{0.28\linewidth}
    \centering
    \vspace{-8mm}
    \begin{subfigure}[t]{\linewidth}
        \centering
        \includegraphics[width=0.8\linewidth]{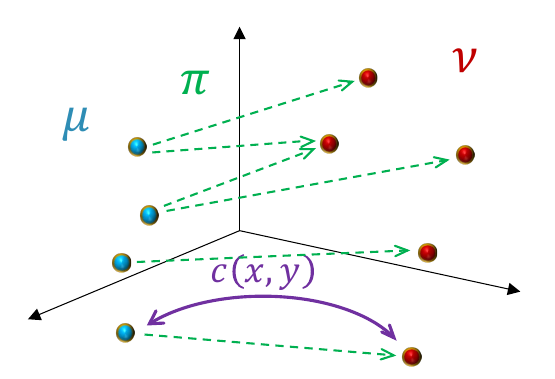}
        \caption{A coupling $\pi\in\Pi(\mu,\nu)$ transports mass from $\mu$ to $\nu$.}\label{fig:OT_geometry}
    \end{subfigure}
    \vspace{1
    mm}

    \begin{subfigure}[t]{\linewidth}
        \centering
        \begin{tikzpicture}[x=0.46cm,y=0.46cm, every node/.style={font=\scriptsize}]
            \def\ncols{5}
            \def\nrows{4}
            \foreach \r/\c/\a in {0/0/12,0/1/26,0/2/78,0/3/18,0/4/6,1/0/8,1/1/58,1/2/22,1/3/14,1/4/10,2/0/42,2/1/12,2/2/16,2/3/55,2/4/20,3/0/5,3/1/10,3/2/14,3/3/24,3/4/80}{
                \fill[green!\a!white] (\c,\nrows-1-\r) rectangle ++(1,1);
            }
            \draw[black!70, thick] (0,0) rectangle (\ncols,\nrows);
            \foreach \x in {1,...,4}{\draw[black!50] (\x,0) -- (\x,\nrows);}
            \foreach \y in {1,...,3}{\draw[black!50] (0,\y) -- (\ncols,\y);}
            \node[text=green!65!black, font=\normalsize] at (5.4,3.8) {$\Pi$};
            \node[text=red!75!black, font=\normalsize] at (2.5,-0.98) {$\nu$};
            \node[text=cyan!70!black, font=\normalsize, rotate=90] at (-1.02,2.0) {$\mu$};
            \foreach \c in {0,...,4}{\node[text=red!75!black] at (\c+0.5,-0.36) {$y_{\the\numexpr\c+1\relax}$};}
            \foreach \r in {0,...,3}{\node[text=cyan!70!black] at (-0.42,\nrows-0.5-\r) {$x_{\the\numexpr\r+1\relax}$};}
        \end{tikzpicture}
        \vspace{-1.4mm}
        \caption{Coupling matrix $\Pi$, representing the correspondence.}\label{fig:OT_coupling}
    \end{subfigure}
    \vspace{-1mm}
    \caption{Optimal transport.}
    \vspace{-4mm}
\end{wrapfigure}

OT compares two probability distributions by the minimum cost of moving mass from one to the other. Given $\mu\in\mathcal{P}(\mathcal{X})$, $\nu\in\mathcal{P}(\mathcal{Y})$, and a cost function $c:\mathcal{X}\times\mathcal{Y}\to\R$, the Kantorovich OT problem is \citep{villani2003topics,peyre2019computational}
$\mathsf{OT}_c(\mu, \nu) \coloneqq \inf_{\pi \in \Pi(\mu,\nu)} \int c(x, y)\, d\pi(x, y)$, where  $\Pi(\mu,\nu)\coloneqq \{\pi\in\mathcal{P}(\mathcal{X}\times \mathcal{Y}):\,\pi(\cdot\times\mathcal{Y})=\mu,\,\pi(\mathcal{X}\times\cdot)=\nu\}$ is the set of all couplings between $\mu$ and $\nu$, each specifying how mass under $\mu$ is matched to mass under $\nu$. The OT coupling can be~viewed as a \emph{soft correspondence}, since the mass associated with one support point of $\,\mu\,$ may be split across multiple support points of $\nu$, with the coupling weights quantifying the association~strength. We use the squared Euclidean cost $c(x,y)=\|x-y\|^2$; see \Cref{fig:OT_geometry}. For discrete measures with $m$ and $n$ support~points, the optimal coupling is a nonnegative matrix $\Pi\in\R_{\ge0}^{m\times n}$, as illustrated in \Cref{fig:OT_coupling}, whose entry $\Pi_{i,j}$ records the mass transported from support point $i$ of $\mu$ to support point $j$ of $\nu$. In PLOT, the coupling matrix will provide the soft correspondence between abstract variables and candidate neural sites.

\paragraph{Entropic OT.} For computation, we use entropically regularized OT
\begin{equation}
\mathsf{OT}_{c}^{\varepsilon} (\mu,\nu) = \inf_{\pi \in \Pi (\mu,\nu)} \int c \, d\pi + \varepsilon \mathsf{KL}(\pi \| \mu \otimes \nu), \label{eq:eot}
\end{equation}
where $\varepsilon>0$. The added Kullback-Leibler (KL) divergence penalty  makes the problem strictly convex and enables efficient computation via Sinkhorn iterations, with $O(mn)$ work per iteration for measures with $m$ and $n$ support points \citep{cuturi2013sinkhorn,peyre2019computational}.


\paragraph{Unbalanced OT.}
Standard OT enforces both marginals exactly, which can be too rigid when some candidate neural sites are irrelevant to the target variables. UOT relaxes the strict marginal constraints by penalizing, rather than forbidding, marginal mismatch~\citep{chizat2018scaling,peyre2019computational}:
\begin{equation}
\mathsf{OT}_{c}^{\varepsilon,\beta_1,\beta_2}(\mu,\nu)
\coloneqq
\inf_{\pi\in\mathcal{M}_+(\mathcal{X}\times\mathcal{Y})}
\int c\,d\pi +\varepsilon\mathsf{KL}(\pi\|\mu\otimes\nu)+\beta_1\mathsf{KL}(\pi_1\|\mu)+\beta_2\mathsf{KL}(\pi_2\|\nu),
\label{eq:uot}
\end{equation}
where $\pi_1$ and $\pi_2$ are the first and second marginals of $\pi$, which need not coincide exactly with $\mu$ and $\nu$ under the UOT framework. Large $\beta_1,\beta_2$ values recover balanced EOT, while finite values allow some mass to remain unmatched. In PLOT, when the neural candidate family is large and may contain irrelevant sites, we use the one-sided version that fixes the abstract marginal $\pi_1=\mu$ and relaxes only the neural marginal $\pi_2$, with parameter $\beta=\beta_2$. This allows the coupling to assign little mass to irrelevant neural sites while retaining all abstract variables in the correspondence.

\section{Progressive Localization via Optimal Transport}\label{sec:PLOT_method}

We now describe PLOT for localizing neural intervention handles associated with abstract variables $Z_1,\ldots,Z_m$, using the setting of \Cref{subsec:background_causal}. Assume the neural model has $L$ layers, whose activation vectors, given $x$, are denoted by $h_\ell(x)\in\mathbb{R}^{d_\ell}$, $\ell=1,\ldots,L$. We define candidate \emph{neural sites} $s_1,\ldots,s_n$, each specifying an intervention subspace within a parent layer. A site may range from an entire activation vector to a coordinate group or PCA span, and the candidate family may span multiple layers. We denote the dimension of $s_j$ by $r_j$, with activations $s_j(x)\in\mathbb{R}^{r_j}$ on input $x$.

Our method requires three datasets of base/source inputs: $\mathcal{D}_{\mathrm{ft}},\mathcal{D}_{\mathrm{cal}},\mathcal{D}_{\mathrm{te}}$, corresponding to fit, calibration, and test banks, which are used, respectively, to construct signatures, select localization/handle parameters, and evaluate the selected handles. We depart from the conventional train/validation/test terminology to better reflect the distinct roles these datasets play in our method. We first delineate a single PLOT stage and then discuss progressive localization.




\subsection{Single PLOT Stage}\label{subsec:PLOT_single}

We first describe the four steps for constructing and evaluating an intervention handle. 



\paragraph{Step 1: Effect signatures.}
We use $\mathcal{D}_{\mathrm{ft}}$ to record the output effects of interchange interventions on each abstract variable $Z_1,\ldots,Z_m$ and neural site $s_1,\ldots,s_n$. These output effects are called \emph{effect signatures}, which are constructed to lie in a common space and serve to align the models via OT.


\vspace{-1mm}
\underline{Causal signatures:}
As in \Cref{subsec:background_causal}, for each pair $(x_t^b,x_t^s)\in\mathcal{D}_{\mathrm{ft}}$ and variable $Z_i$, write $y^{\mathrm{abs\_swap}}_{i,t}$ for the abstract interchange output and $y_t^{\mathrm{abs}}=\mathfrak{C}(x_t^b)$ for the factual output. The \textit{causal effect signature} is
\[
\Delta_{i,t}^{\mathrm{abs}}
\coloneqq
d^\mathrm{abs}\bigl(y_{i,t}^{\mathrm{abs\_swap}},y_t^{\mathrm{abs}}\bigr),
\qquad
i=1,\ldots,m,\quad t=1,\ldots,T_{\mathrm{ft}},
\]
where we typically take $d^\mathrm{abs}(y,y')=y-y'$ in the chosen numerical encoding of the outputs.


\underline{Neural signatures:}
Let $\ell(j)\in\{1,\ldots,L\}$ denote the parent layer of site $s_j$, so that $s_j$ can be viewed as an $r_j$-dimensional intervention subspace of the layer activation space $\mathbb{R}^{d_{\ell(j)}}$. Let $W_j\in\mathbb{R}^{d_{\ell(j)}\times r_j}$ with $W_j^\intercal W_j=I_{r_j}$ have columns forming an orthonormal basis for this subspace. Thus, $W_j^\intercal$ maps the full layer activation $h_{\ell(j)}(x)$ to its coordinates within site $s_j$, while $W_j$ maps changes in these coordinates back into the full layer activation space. We thus have $s_j(x)\coloneqq W_j^\intercal h_{\ell(j)}(x)\in\mathbb{R}^{r_j}$. In the canonical basis, the columns of $W_j$ are canonical basis vectors selecting an individual neuron or coordinate block. In the PCA basis, they are PCA directions fitted to the layer's fit-bank activations.


Given $(x_t^b,x_t^s)$, the neural swap/intervention at $s_j$ replaces the base coordinates in this subspace by the corresponding source coordinates, while leaving the orthogonal complement unchanged:
\[
h_{\ell(j)}
\leftarrow
h_{\ell(j)}(x_t^b)
+
W_j\left(
s_j(x_t^s)-s_j(x_t^b)
\right)=
h_{\ell(j)}(x_t^b)
+
W_jW_j^\intercal
\left(
h_{\ell(j)}(x_t^s)-h_{\ell(j)}(x_t^b)
\right).
\]

Writing the output as $y^{\mathrm{nn\_swap}}_{j,t}$ and the factual output as $y_t^{\mathrm{nn}}\coloneqq \mathfrak{N}(x_t^b)$, the \emph{neural effect~signature} is
\[
    \Delta_{j,t}^{\mathrm{nn}}
    \coloneqq
    d^\mathrm{nn}\big(y_{j,t}^{\mathrm{nn\_swap}},y_{t}^{\mathrm{nn}}\big),
    \qquad
    j=1,\ldots,n,\quad t=1,\ldots,T_{\mathrm{ft}}.
\]
We usually take $d^\mathrm{nn}(y,y') = \mathrm{softmax}\big(y)-\mathrm{softmax}\big(y')$, with other options including summaries based on the total variation distance $d^\mathrm{nn}(y,y') = \frac{1}{2} \left\|
\operatorname{softmax}\!\left(y\right)
-
\operatorname{softmax}\!\left(y'\right)
\right\|_1$, KL divergence, or any other summary of the neural base and swap outputs. 
See \Cref{app:ablations-signatures} for an empirical comparison of several such effect signatures.


\paragraph{Step 2: Optimal transport matching.}
We associate each abstract variable and neural site with an effect signature, obtained by aggregating its output effects over the fit bank. We then fit an OT coupling between the signature collections, using UOT when irrelevant neural sites may be present. The coupling yields a weighted soft correspondence between abstract variables and neural sites.


Specifically, we pool effect signatures over fit pairs to obtain one signature per~variable/site:
\[
    u_i\coloneqq \operatorname{Agg}\!\left(\Delta^{\mathrm{abs}}_{i,1},\ldots,\Delta^{\mathrm{abs}}_{i,T_{\mathrm{ft}}}\right),
    \qquad
    v_j\coloneqq \operatorname{Agg}\!\left(\Delta^{\mathrm{nn}}_{j,1},\ldots,\Delta^{\mathrm{nn}}_{j,T_{\mathrm{ft}}}\right).
\]
For example, $\operatorname{Agg}$ can average or concatenate the per-pair effects, optionally followed by $\ell_2$-normalization. These signatures define empirical measures $\mu\coloneqq \frac{1}{m}\sum_{i=1}^m \delta_{u_i}$ and $\nu\coloneqq \frac{1}{n}\sum_{j=1}^n \delta_{v_j}$, where $\delta_x$ is a Dirac mass. We compute the EOT coupling $\Pi^\star_\varepsilon\in\mathbb{R}_{\ge 0}^{m\times n}$ between $\mu$ and $\nu$ via Sinkhorn's algorithm with quadratic cost $c(u_i,v_j)=\|u_i-v_j\|^2$ \citep{cuturi2013sinkhorn}. For UOT, we use the one-sided specialization described above, fixing $\pi_1=\mu$ and relaxing only $\pi_2$, with parameter $\beta=\beta_2$. Slightly abusing notation, write $\Pi$ for the fitted EOT or UOT coupling, whose $i$-th row $[\Pi]_{i:}\in\mathbb{R}^{n}$ gives the matching weights of $Z_i$ to $s_1,\ldots,s_n$.


\paragraph{Step 3: Handle calibration.}
Next, we use the fitted coupling $\Pi$ and $\mathcal{D}_{\mathrm{cal}}$ to calibrate an \emph{intervention handle} for each abstract variable. A handle combines interventions at selected neural sites, with relative strengths determined by $\Pi$.\footnote{In progressive localization, intermediate stages use calibration data to select parent sites, while the final stage calibrates the intervention handle.} For abstract variable $Z_i$, calibration selects the number of neural sites $K$ and a global intervention strength $\lambda$. Let $\mathcal{S}_i^K$ denote the indices of the $K$ largest entries in the $i$-th row of $\Pi$. We restrict that row to $\mathcal{S}_i^K$ and renormalize $\widetilde{\Pi}_{j\mid i}^{K}\coloneqq\Pi_{ij}\mathds{1}_{\{j\in\mathcal{S}_i^K\}}/\big(\sum_{j'\in\mathcal{S}_i^K}\Pi_{ij'}\big)$, for $j=1,\ldots,n$. Suppose first that the selected sites lie in one activation vector $h(x)\in\mathbb{R}^d$. Applied to a base/source pair $(x_t^b,x_t^s)\in\mathcal{D}_{\mathrm{cal}}$, the corresponding handle performs
\[
h
\leftarrow
h(x_t^b)
+
\lambda
\sum_{j\in\mathcal{S}_i^K}
\widetilde{\Pi}_{j\mid i}^{K}
W_j\!\left(s_j(x_t^s)-s_j(x_t^b)\right).
\]
When the selected sites lie in multiple layers, we apply the interventions in forward-pass order. At each later layer, the update is applied to the activation produced by the already-intervened computation.


Let $H_i^{K,\lambda}$ denote this candidate handle and let
$y_{i,t}^{\mathrm{nn\_swap}}(K,\lambda)$ be the neural output obtained by applying it to calibration pair $(x_t^b,x_t^s)$. We evaluate it using
\begin{equation}\label{EQ:IIA_cal}
\operatorname{IIA}_{\mathrm{cal}}(i;K,\lambda)
\coloneqq
\frac{1}{T_{\mathrm{cal}}}
\sum_{t=1}^{T_{\mathrm{cal}}}
\mathds{1}_{\left\{
y_{i,t}^{\mathrm{nn\_swap}}(K,\lambda)
\sim
y_{i,t}^{\mathrm{abs\_swap}}
\right\}},
\end{equation}
where $\sim$ denotes task-dependent agreement between the neural and abstract intervention outputs. For an LLM, agreement holds when the highest-logit intervened token equals the abstract target. We select
$(K_i^\star,\lambda_i^\star)\in
\operatorname*{arg\,max}_{K\in\mathcal K,\;\lambda\in\Lambda}
\operatorname{IIA}_{\mathrm{cal}}(i;K,\lambda)$
and set the calibrated handle as $H_i^\star\coloneqq H_i^{K_i^\star,\lambda_i^\star}$. We repeat this for each abstract variable, equivalently, for each coupling row. EOT/UOT parameters and site resolutions, when searched, are also selected on $\mathcal D_{\mathrm{cal}}$.

\paragraph{Step 4: Handle evaluation.}
Finally, we evaluate the IIA score, similarly to \eqref{EQ:IIA_cal}, for each calibrated intervention handle on the holdout test set $\mathcal{D}_{\mathrm{te}}$, without further parameter selection. Unless stated otherwise, the IIA reported in our experiments is the average test IIA over all abstract variables.

\begin{remark}[Alternative matching]
OT provides a principled matching framework, which we adopt in PLOT. We note that other constructions of $\Pi$ are possible, e.g., based on cosine similarities between effect signatures or direct IIA matching. We compare these alternatives with OT in \Cref{app:ablations-matching}.
\end{remark}


\medskip

\subsection{Multi-Stage Progressive Localization}\label{subsec:progressive_localization}

PLOT becomes hierarchical by composing Steps 1--2 across nested neural site families. Rather than matching all sites at the finest resolution at once, progressive localization first selects coarse parent sites, then fits new couplings within them. The number of stages depends on the architecture and desired granularity, as illustrated in \Cref{fig:PLOT_diagram}. The high-level causal model remains fixed throughout. The progressive stages refine neural resolution rather than define distinct levels of causal abstraction.

At stage $r$, let $\mathcal{T}_i^{(r)}$ denote the current candidate family for localizing $Z_i$. PLOT fits a coupling $\Pi_i^{(r)}$ between all target variables $Z_1,\ldots,Z_m$ and the sites in $\smash{\mathcal{T}_i^{(r)}}$, using OT or one-sided UOT as in Step 2. The first stage uses a common candidate family and coupling for all variables. At each intermediate stage $r$, we take the highest-mass sites in the row $\smash{[\Pi_i^{(r)}]_{i:}}$, evaluate their individual interventions on $\mathcal{D}_{\mathrm{cal}}$, and select the site with the highest calibration IIA for further refinement. Subdividing that parent yields the finer candidate family $\smash{\mathcal{T}_i^{(r+1)}}$ for the next stage. At the final stage $r_\text{final}$, Step 3 calibrates the $i$-th row of $\Pi_i^{(r_\text{final})}$ into an intervention handle for $Z_i$, and Step 4 evaluates it on $\mathcal{D}_{\mathrm{te}}$. 
The final localization may also guide DAS, where PLOT selects coarse neural locations, such as a token position and layer, and then DAS learns the intervention subspace within that layer.

\section{Experimental Settings and Results}
\label{sec:experiments_results}


Our experiments are organized by increasing localization complexity:
(i) HEQ, where we test single-stage handle construction over individual neurons;
(ii) binary addition, where we compare global and progressive localization of carry variables in native and PCA representations; and
(iii) MCQA, where we examine accuracy under matched handle-dimension limits, runtime, and learned-handle structure, test UOT for broad candidate families with irrelevant sites, and compare against masking, PLOT-guided DAS, and full DAS. See \Cref{app:tasks} for full experiment details.


\subsection{Hierarchical Equality (HEQ)}
\label{subsec:current-heq-results}

\begin{wrapfigure}{r}{0.36\linewidth}
    \centering
    \vspace{-5.6mm}
    \begin{tikzpicture}[>=Latex, scale=0.82, transform shape, every node/.style={font=\scriptsize}, line cap=round, line join=round]
        \node[inputvar, minimum size=7.1mm, inner sep=0.7pt] (w) at (-1.45,2.05) {$W$};
        \node[inputvar, minimum size=7.1mm, inner sep=0.7pt] (x) at (0.45,2.05) {$X$};
        \node[inputvar, minimum size=7.1mm, inner sep=0.7pt] (y) at (2.05,2.05) {$Y$};
        \node[inputvar, minimum size=7.1mm, inner sep=0.7pt] (z) at (3.95,2.05) {$Z$};
        \node[targetvar, minimum size=7.1mm, inner sep=0.7pt] (wx) at (-0.55,1.15) {$z_{WX}$};
        \node[targetvar, minimum size=7.1mm, inner sep=0.7pt] (yz) at (3.05,1.15) {$z_{YZ}$};
        \node[outnode, minimum size=7.1mm, inner sep=0.7pt] (o) at (1.25,0.25) {$y$};

        \draw[->] (w) -- (wx);
        \draw[->] (x) -- (wx);
        \draw[->] (y) -- (yz);
        \draw[->] (z) -- (yz);
        \draw[->] (wx) -- (o);
        \draw[->] (yz) -- (o);
    \end{tikzpicture}
    \caption{Causal model for HEQ.} 
    \label{fig:heq-causal-model}
     \vspace{-5mm}
\end{wrapfigure}

As a first single-stage test of PLOT, we consider the HEQ benchmark from \citet{geiger2024das}, where the goal is to determine whether two pairs of objects have the same equality relation. On this small task, a single PLOT matching over all hidden neurons yields accurate, compact handles without progressive localization or learned intervention rotations.


\paragraph{Causal model.} As in \Cref{fig:heq-causal-model}, inputs $W,X,Y,Z\in[100]$ define equality variables $z_{WX}=\mathds{1}_{\{W=X\}}$ and $z_{YZ}=\mathds{1}_{\{Y=Z\}}$. The output $y=\mathfrak{C}(W,X,Y,Z)=\mathds{1}_{\{z_{WX}=z_{YZ}\}}$ is $1$ precisely when both pairs are equal or both are unequal. For example, $\mathfrak{C}(1,1,2,2)=\mathfrak{C}(1,2,3,4)=1$, while $\mathfrak{C}(1,1,1,2)=0$.


\paragraph{Neural model.}
We encode each of $W,X,Y,Z$ with a fixed 4-dimensional embedding and concatenate them into an input $x\in\mathbb{R}^{16}$. The neural model is a ReLU MLP $\mathfrak{N}:\mathbb{R}^{16}\to\mathbb{R}^2$ with three hidden layers of width $16$, trained to $100\%$ accuracy on held-out factual data. See \Cref{app:tasks-heq} for details.




\paragraph{Single-stage PLOT.}
We use disjoint fit, calibration, and test pair banks; see \Cref{app:tasks-heq} for their construction. The abstract variables are $Z_1=z_{WX}$ and $Z_2=z_{YZ}$, while the candidate sites are all $3\times16=48$ individual hidden neurons. For the effect signatures, we encode abstract outputs as one-hot vectors and take $d^{\mathrm{abs}}(y,y')=y-y'$, while for neural logits we take $d^{\mathrm{nn}}(y,y')=\mathrm{softmax}(y)-\mathrm{softmax}(y')$, placing both signatures in the same $\mathbb{R}^2$ space. We fit the resulting $2\times48$ EOT coupling on $\mathcal{D}_{\mathrm{ft}}$ using Sinkhorn's algorithm and calibrate the entropic parameter $\varepsilon\in\{0.01,0.25,2,4,8,16\}$, support size $K\in\{1,\ldots,16\}$, and intervention strength $\lambda\in\{1,\ldots,16\}$ on $\mathcal{D}_{\mathrm{cal}}$. As a baseline, DAS trains rotations on $\mathcal{D}_{\mathrm{ft}}$ for each variable over layers $\ell=1,2,3$ and subspace dimensions $r\in\{1,\ldots,16\}$, and selects the best handle on $\mathcal{D}_{\mathrm{cal}}$.

\begin{figure}[t]
    \centering
    \newlength{\heqresultsheight}
    \setlength{\heqresultsheight}{0.27\linewidth}
    \begin{subfigure}[t]{0.58\linewidth}
        \centering
        \begin{minipage}[c][\heqresultsheight][c]{\linewidth}
        \centering
        \scriptsize
        \setlength{\tabcolsep}{4.0pt}
        \renewcommand{\arraystretch}{1.08}
        \resizebox{\linewidth}{!}{%
        \begin{tabular}{lcc}
        \toprule
        Metric & \PLOT{} & DAS \\
        \midrule
        $z_{WX}$ sensitivity & $0.9923 \pm 0.0155$ & $0.9938 \pm 0.0041$ \\
        $z_{WX}$ invariance  & $0.9988 \pm 0.0026$ & $0.9998 \pm 0.0004$ \\
        $z_{YZ}$ sensitivity & $0.9913 \pm 0.0190$ & $0.9934 \pm 0.0047$ \\
        $z_{YZ}$ invariance  & $0.9901 \pm 0.0209$ & $0.9999 \pm 0.0003$ \\
        \midrule
        Mean sensitive IIA   & $0.9918 \pm 0.0172$ & $0.9936 \pm 0.0032$ \\
        Avg. IIA             & $0.9931 \pm 0.0127$ & $0.9967 \pm 0.0016$ \\
        Serial runtime (s)   & $3.94 \pm 0.27$     & $184.31 \pm 44.41$ \\
        \bottomrule
        \end{tabular}
        }
        \end{minipage}
        \vspace{-1.5mm}
        \caption{Accuracy and serial runtime summary.}
        \label{fig:heq-joint-table}
    \end{subfigure}\hfill
    \begin{subfigure}[t]{0.40\linewidth}
        \centering
        \begin{minipage}[c][\heqresultsheight][c]{\linewidth}
        \centering
        \includegraphics[width=\linewidth]{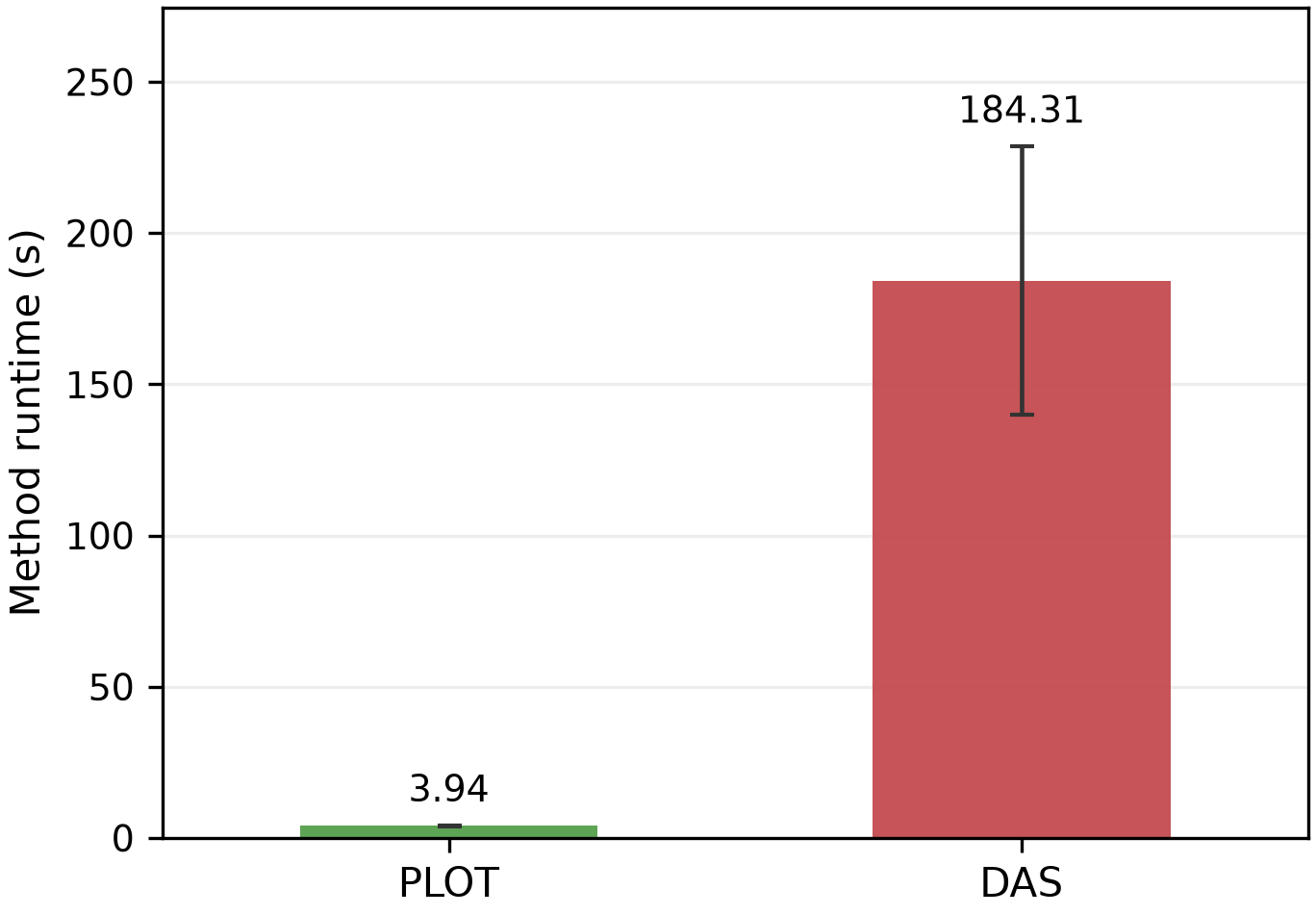}
        \end{minipage}
        \vspace{-2mm}
        \caption{Serial runtime.}
        \label{fig:heq-joint-runtime}
    \end{subfigure}
    \caption{HEQ comparison over 10 seeds. Values are mean $\pm$ standard deviation. Runtime excludes neural model training and includes fitting, calibration, and test evaluation.}   
    \vspace{-2.5mm}
    \label{fig:heq-joint-results}
\end{figure}

\begin{figure}[t]
    \centering
    \includegraphics[width=\linewidth]{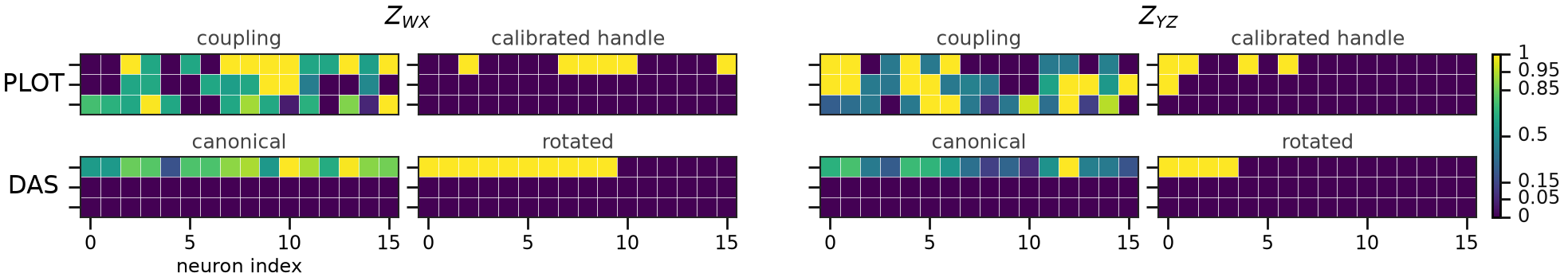}
    \vspace{-7mm}
    \caption{HEQ matching weights and intervention handles for one illustrative seed (seed~7). PLOT is shown before and after top-$K$ calibration, and DAS in the canonical and rotated representations.} 
    \label{fig:heq-handle-heatmaps}
    \vspace{-3.5mm}
\end{figure}


\paragraph{Results and learned handles.}

The table in \Cref{fig:heq-joint-results} reports test IIA over seeds~1--10 for $z_{WX}$ and $z_{YZ}$, together with their average. For each abstract variable, we separately evaluate sensitivity, on pairs that change that variable, and invariance, on pairs that leave it fixed. Both single-stage PLOT and DAS recover accurate handles on this benchmark, with average test IIA $0.9931$ and $0.9967$, respectively. Thus, PLOT obtains high interchange intervention agreement directly over the original hidden neurons without learning a rotation. PLOT takes $3.94$ seconds on average, compared with $184.31$ seconds for the DAS layer and dimension search, a speedup of approximately $47\times$. Runtime includes effect signature construction, coupling fitting or rotation training, calibration, and held-out evaluation, but excludes neural model training. See~\Cref{app:tasks-heq} for hardware and timing details.


A structural difference between the methods is shown in \Cref{fig:heq-handle-heatmaps} for seed~7. DAS is fit layer by layer, whereas PLOT fits one transport plan over all candidate neurons simultaneously. Calibration converts the resulting matching weights into a compact top-$K$ intervention handle. In contrast, DAS reaches similar performance through a learned rotated subspace. Further layerwise summaries, OT parameter sensitivity, and learned intervention sizes are reported in \Cref{app:tasks-heq}.

\subsection{Binary Addition}\label{subsec:binary-addition}

\begin{wrapfigure}{r}{0.56\linewidth}
    \centering
    \vspace{-10mm}
    \begin{tikzpicture}[>=Latex, every node/.style={font=\scriptsize}, line cap=round, line join=round, scale=0.7, transform shape]
        \node[inputvar, minimum size=6.1mm, inner sep=1.4pt] (a0) at (0.00,3.3) {$A_0$};
        \node[inputvar, minimum size=6.1mm, inner sep=1.4pt] (b0) at (0.90,3.3) {$B_0$};
        \node[scmnode, minimum size=6.1mm, inner sep=1.4pt] (s0) at (0.0,1.8) {$S_0$};
        \node[targetvar, minimum size=6.1mm, inner sep=1.4pt] (c1) at (0.90,1.8) {$C_1$};

        \node[scmnode, minimum size=6.1mm, inner sep=1.4pt] (c1top) at (2.00,3.3) {$C_1$};
        \node[inputvar, minimum size=6.1mm, inner sep=1.4pt] (a1) at (2.90,3.3) {$A_1$};
        \node[inputvar, minimum size=6.1mm, inner sep=1.4pt] (b1) at (3.80,3.3) {$B_1$};
        \node[scmnode, minimum size=6.1mm, inner sep=1.4pt] (s1) at (2.90,1.8) {$S_1$};
        \node[scmnode, minimum size=6.1mm, inner sep=1.4pt] (c2) at (3.80,1.8) {$C_2$};
        
        \node[scmnode, minimum size=6.1mm, inner sep=1.4pt] (c2top) at (4.90,3.3) {$C_2$};
        \node[inputvar, minimum size=6.1mm, inner sep=1.4pt] (a2) at (5.80,3.3) {$A_2$};
        \node[inputvar, minimum size=6.1mm, inner sep=1.4pt] (b2) at (6.70,3.3) {$B_2$};
        \node[scmnode, minimum size=6.1mm, inner sep=1.4pt] (s2) at (5.80,1.8) {$S_2$};
        \node[scmnode, minimum size=6.1mm, inner sep=1.4pt] (c3) at (6.70,1.8) {$C_3$};

        \node[scmnode, minimum size=6.1mm, inner sep=1.4pt] (c3top) at (7.80,3.3) {$C_3$};
        \node[inputvar, minimum size=6.1mm, inner sep=1.4pt] (a3) at (8.70,3.3) {$A_3$};
        \node[inputvar, minimum size=6.1mm, inner sep=1.4pt] (b3) at (9.60,3.3) {$B_3$};
        \node[scmnode, minimum size=6.1mm, inner sep=1.4pt] (s3) at (8.70,1.8) {$S_3$};
        \node[scmnode, minimum size=6.1mm, inner sep=1.4pt] (c4) at (9.60,1.8) {$C_4$};

        \node[outnode, minimum size=6.1mm, inner sep=1.4pt] (y) at (4.85,0.10) {$y$};
        \node[text=red!70!black, below=2mm of y] {$y=(C_4,S_3,S_2,S_1,S_0)$};

        \draw[->] (a0) -- (s0);
        \draw[->] (b0) -- (s0);
        \draw[->] (a0) -- (c1);
        \draw[->] (b0) -- (c1);

        \draw[->] (c1) to[out=15,in=165] (c1top);
        \draw[->] (a1) -- (s1);
        \draw[->] (b1) -- (s1);
        \draw[->] (c1top) -- (s1);
        \draw[->] (a1) -- (c2);
        \draw[->] (b1) -- (c2);
        \draw[->] (c1top) -- (c2);

        \draw[->] (c2) to[out=15,in=165] (c2top);
        \draw[->] (a2) -- (s2);
        \draw[->] (b2) -- (s2);
        \draw[->] (c2top) -- (s2);
        \draw[->] (a2) -- (c3);
        \draw[->] (b2) -- (c3);
        \draw[->] (c2top) -- (c3);

        \draw[->] (c3) to[out=15,in=165] (c3top);
        \draw[->] (a3) -- (s3);
        \draw[->] (b3) -- (s3);
        \draw[->] (c3top) -- (s3);
        \draw[->] (a3) -- (c4);
        \draw[->] (b3) -- (c4);
        \draw[->] (c3top) -- (c4);

        \draw[->] (s0) -- (y.180);
        \draw[->] (s1) -- (y.135);
        \draw[->] (s2) -- (y.90);
        \draw[->] (s3) -- (y.45);
        \draw[->] (c4) -- (y.0);

    \end{tikzpicture}
    \vspace{-2mm}
    \caption{Causal model for $4$-bit ripple-carry addition.}
    \label{fig:binary-addition-causal}
    \vspace{-5mm}
\end{wrapfigure}

We next consider binary addition with a GRUCell-based recurrent network, where we seek to locate internal carry variables across timesteps and within recurrent states. We compare global and progressive localization in native and PCA coordinates, and show how PLOT can guide DAS without searching every timestep.

\paragraph{Causal model.}
As in \Cref{fig:binary-addition-causal}, the inputs are two $4$-bit numbers $A_3A_2A_1A_0$ and $B_3B_2B_1B_0$. The abstract model is the standard ripple-carry adder, with $S_0=(A_0+B_0)\bmod 2$, $C_1=\lfloor(A_0+B_0)/2\rfloor$, and for $i=1,2,3$, $S_i=(A_i+B_i+C_i)\bmod 2$ and $C_{i+1}=\lfloor(A_i+B_i+C_i)/2\rfloor$. The output is $y=(C_4,S_3,S_2,S_1,S_0)$. We report intervention accuracy for the internal carry variables $C_1,C_2,C_3$.

\paragraph{Neural model.}
The neural model reads bit-pairs $(A_\ell,B_\ell)$ from least to most significant and updates a GRUCell state $h_\ell\in\mathbb{R}^{16}$ at timestep $\ell$, for $\ell=0,1,2,3$. Linear readouts predict the five output bits $(C_4,S_3,S_2,S_1,S_0)$. The model is trained to solve $4$-bit addition exactly; architecture and training details are in \Cref{app:tasks-addition}. Thus, $h_\ell$ is the natural coarse neural site for carry $C_{\ell+1}$.


\paragraph{Global and progressive localization.}
We use the output-effect pipeline of \Cref{subsec:PLOT_single} with binary-addition-specific pair banks and recurrent candidate sites. Following the MIB Arithmetic benchmark \citep{mueller2025mib}, we construct structured base/source banks with operand flips and carry-targeting sources, split into fit, calibration, and test banks; see \Cref{app:tasks-addition}. Candidate sites~are native-coordinate blocks or PCA spans, with PCA fitted to fit-bank hidden activations at each timestep.

For each resolution, one-stage PLOT fits one coupling between the three carries and sites across all four timesteps, allowing handles to span timesteps. Two-stage PLOT first fits a $3\times4$ coupling over the recurrent states to select one timestep per carry, then fits new couplings over finer sites within the selected states. Both use EOT, with $\varepsilon$, site resolution $r$, support size $K$, and strength $\lambda$ selected on $\mathcal{D}_{\mathrm{cal}}$. PLOT-guided DAS (PLOT-DAS) uses the coarse localization to train DAS only within selected states, while full DAS trains rotations at every timestep. Both select final handles on $\mathcal{D}_{\mathrm{cal}}$.

\begin{figure}[!t]
    \centering
    \newlength{\binaryadditionresultsheight}
    \setlength{\binaryadditionresultsheight}{0.27\linewidth}

    \begin{subfigure}[t]{0.58\linewidth}
        \centering
        \begin{minipage}[c][\binaryadditionresultsheight][c]{\linewidth}
        \centering
        \scriptsize
        \setlength{\tabcolsep}{4.0pt}
        \renewcommand{\arraystretch}{1.08}
        \resizebox{\linewidth}{!}{%
        \begin{tabular}{lcc}
            \toprule
            Method & Test IIA & Serial runtime (s) \\
            \midrule

            \onePLOTnat{}
            & $0.767 \pm 0.032$
            & $24.6 \pm 3.7$ \\

            \onePLOTpca{}
            & $0.900 \pm 0.053$
            & $24.7 \pm 2.5$ \\

            \twoPLOTnat{}
            & $0.792 \pm 0.023$
            & $20.1 \pm 2.1$ \\

            \twoPLOTpca{}
            & $\mathbf{0.936 \pm 0.028}$
            & $20.4 \pm 1.4$ \\

            \midrule

            \PLOTdas{}
            & $\mathbf{0.991 \pm 0.007}$
            & $\mathbf{15.2 \pm 0.8}$ \\

            Full DAS
            & $\mathbf{0.991 \pm 0.007}$
            & $54.7 \pm 2.1$ \\

            \bottomrule
        \end{tabular}
        }
        \end{minipage}
        \vspace{-2mm}
        \caption{Accuracy and runtime summary.}
        \label{fig:binary-addition-guided-table}
    \end{subfigure}
    \hfill
    \begin{subfigure}[t]{0.40\linewidth}
        \centering
        \begin{minipage}[c][\binaryadditionresultsheight][c]{\linewidth}
        \centering
\includegraphics[width=\linewidth]{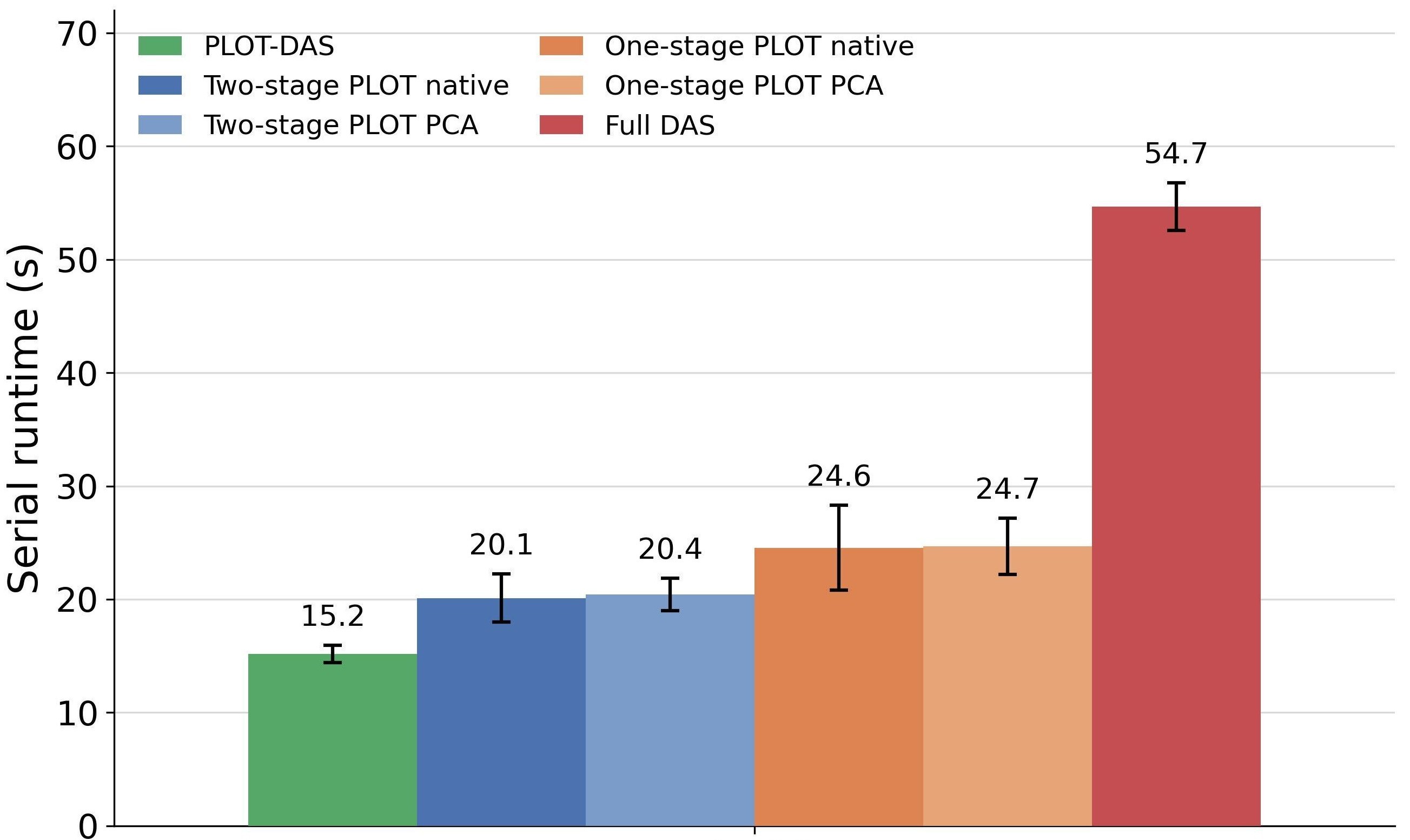}
        \end{minipage}
        \vspace{-2mm}
        \caption{Serial runtime.}
        \label{fig:binary-addition-guided-runtime}
    \end{subfigure}
    \vspace{-2mm}
    \caption{Binary-addition results over five seeds. Test IIA equally weights sensitivity and invariance and averages over $C_1,C_2,C_3$. Runtime covers the full parameter~grids.}
    \label{fig:binary-addition-guided-results}
    \vspace{-3mm}
\end{figure}

\paragraph{Results and learned handles.}
\Cref{fig:binary-addition-guided-results} compares test IIA and serial runtime over five seeds. Moving from native coordinates to PCA substantially improves direct PLOT under both search strategies, from $0.767$ to $0.900$ for one-stage PLOT and from $0.792$ to $0.936$ for two-stage PLOT. Two-stage localization also attains higher mean IIA than one-stage localization in both representations, while reducing runtime from about $24.6$ seconds to $20.1$--$20.4$ seconds. The strongest PLOT-only result is two-stage PLOT PCA, with test IIA $0.936\pm0.028$. PLOT-DAS reaches the same mean test IIA as full DAS, $0.991\pm0.007$, while reducing runtime from $54.7$ to $15.2$ seconds, a roughly $3.6\times$ speedup.

\begin{figure}[t]
    \centering
    \includegraphics[width=\linewidth]{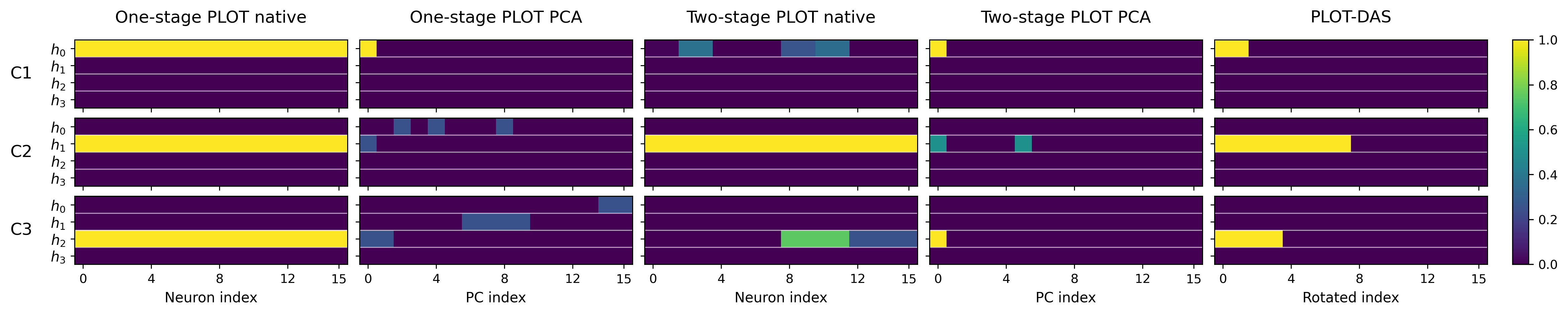}
    \vspace{-5mm}
    \caption{Binary-addition handles for one seed. Rows show carries and columns show methods.} 

    \label{fig:binary-addition-handle-heatmaps}
    \vspace{-5mm}
\end{figure}

\Cref{fig:binary-addition-handle-heatmaps} visualizes the learned handles for one illustrative run (seed~3). One-stage PLOT native selects the full $16$-dimensional states $h_0,h_1,h_2$ for $C_1,C_2,C_3$, while one-stage PLOT PCA yields substantially smaller handles and can combine sites across timesteps. Two-stage PLOT first correctly localizes $C_1,C_2,C_3$ to $h_0,h_1,h_2$, respectively, and the PCA refinement yields compact handles of dimensions $1,2,1$, compared with $8,16,8$ in the native basis. PLOT-DAS selects the same timesteps but learns rotated subspaces of dimensions $2,8,4$. Full DAS is not shown, as it learns the exact same handles as PLOT-DAS through its exhaustive timestep sweep.

\subsection{Multiple-Choice Question Answering (MCQA)}
\label{subsec:mcqa}


\begin{wrapfigure}{r}{0.4\linewidth}
    \centering
     \vspace{-7.5mm}

    \begin{tikzpicture}[>=Latex, scale=0.75, transform shape, every node/.style={font=\scriptsize}, line cap=round, line join=round]
        \tikzset{
            mcqnode/.style={minimum size=7.6mm, inner sep=0.4pt},
            mcqlabel/.style={font=\scriptsize, text=black!65}
        }

        \node[inputvar, mcqnode] (c1) at (-0.2,2.70) {$\mathrm{Col}_1$};
        \node[inputvar, mcqnode] (c2) at (1.15,2.70) {$\mathrm{Col}_2$};
        \node[inputvar, mcqnode] (q)  at (2.50,3.42) {$x$};
        \node[inputvar, mcqnode] (c3) at (3.85,2.70) {$\mathrm{Col}_3$};
        \node[inputvar, mcqnode] (c4) at (5.20,2.70) {$\mathrm{Col}_4$};

        \node[targetvar, mcqnode] (p) at (2.50,1.55) {$\mathrm{AP}$};

        \node[inputvar, mcqnode] (l1) at (-0.2,0.83) {$\mathrm{Sym}_1$};
        \node[inputvar, mcqnode] (l2) at (1.15,0.83) {$\mathrm{Sym}_2$};
        \node[inputvar, mcqnode] (l3) at (3.85,0.83) {$\mathrm{Sym}_3$};
        \node[inputvar, mcqnode] (l4) at (5.20,0.83) {$\mathrm{Sym}_4$};

        \node[outnode, mcqnode] (a) at (2.50,-0.32) {$\mathrm{AT}$};

        \draw[->] (c1) -- (p);
        \draw[->] (c2) -- (p);
        \draw[->] (q) -- (p);
        \draw[->] (c3) -- (p);
        \draw[->] (c4) -- (p);

        \draw[->] (p) -- (a);
        \draw[->] (l1) -- (a);
        \draw[->] (l2) -- (a);
        \draw[->] (l3) -- (a);
        \draw[->] (l4) -- (a);

    \end{tikzpicture}
    \vspace{-1mm}
    \caption[Causal model for MCQA]{MCQA causal model, with abstract variables $\mathrm{AP}$ and $\mathrm{AT}$.}
    \label{fig:mcqa-causal-model}
    \vspace{-1mm}
\end{wrapfigure}

We next evaluate PLOT in the substantially larger setting of the MCQA benchmark from MIB \citep{mueller2025mib}, where Gemma-2-2B answers natural-language multiple-choice prompts. The model reads a color fact and query, identifies the answer containing the queried color, and outputs its symbol. For example, given \emph{``The sky is blue. What color is the sky? (A) red (B) blue (C) green (D) yellow''}, the correct output is \texttt{B}. We compare one-stage and two-stage localization under a handle-dimension limit, together with masking, PLOT-DAS, and full DAS baselines.



\paragraph{Causal model.}
Each input is a natural-language prompt $x$ containing a color fact, a query, and four answer choices. Let $\mathrm{Col}_i$ denote the color at choice position $i$, and $\mathrm{Sym}_i$ be the corresponding answer symbol, e.g., $A,B,C,D$. The abstract model computes the queried color $q(x)$, the answer pointer $\mathrm{AP}=\operatorname*{arg\,unique}_{i\in\{1,2,3,4\}}\{\mathrm{Col}_i=q(x)\}$, and the answer token $\mathrm{AT}=\mathrm{Sym}_{\mathrm{AP}}$; see \Cref{fig:mcqa-causal-model}. The abstract variables of interest are $\mathrm{AP}$ and $\mathrm{AT}$, corresponding to \emph{AnswerPointer} and \emph{Answer} in MIB \citep{mueller2025mib}. We evaluate counterfactuals that change AP, AT, or both; see \Cref{appen:mcqa}.


\paragraph{Neural model.}
We use Gemma-2-2B \citep{gemma2}, a decoder-only transformer with $26$ residual-stream layers and width $2304$. It receives the full MCQA prompt $x$ and is evaluated by its next-token distribution over answer symbols. Localization is performed over final-token activations.



\paragraph{Global and progressive localization.}

We follow the pipeline of \Cref{subsec:binary-addition}, replacing recurrent timesteps with final-token activations across $26$ layers and using the MIB MCQA counterfactual families for fit, calibration, and test. One-stage and two-stage PLOT again use native-coordinate blocks or PCA spans. At the coarse stage, two abstract variables are matched to $26$ layers, some of which may be irrelevant, so we use one-sided UOT to relax the neural marginal. \Cref{tab:mcqa-ot-uot-selected-layers} in the appendix shows consistently stronger UOT layer selection than OT, particularly for $\mathrm{AP}$. Following MIB \citep{mueller2025mib}, we limit each direct handle to $1152$ dimensions, half the residual-stream width, summed across selected layers. We also compare PLOT-DAS, full DAS, and DBM in canonical, PCA, and SAE representations \citep{chaudhary2024dbm}. See \Cref{appen:mcqa} for more details.


\paragraph{Results and learned handles.}

\Cref{tab:mcqa-results} reports test IIA and serial runtime over five seeds. Two-stage PLOT improves mean IIA in both representations, from $0.5955$ to $0.8330$ in native coordinates and from $0.8380$ to $0.8610$ in PCA. For AP, native IIA rises from $0.429$ to $0.763$, while with PCA AP remains at $0.764$ and AT increases from $0.912$ to $0.958$. Two-stage localization is also roughly $2\times$ faster than one-stage matching. The gap is larger than in binary addition due to the larger neural model and candidate family. PCA outperforms native coordinates under both strategies, with two-stage PLOT PCA giving the strongest direct PLOT result with the default signature, consistent with the binary-addition findings. An additional KL-signature ablation yields comparable or higher two-stage IIA with lower observed cross-seed variability but substantially weaker one-stage results; see \Cref{app:ablations-signatures}. DBM PCA is competitive at $0.8570$, whereas the SAE representation considered in MIB is substantially weaker here at $0.7560$. Full DAS remains the most accurate method, while PLOT-DAS retains most of this accuracy at $0.8750$ and cuts runtime to $25.27$ minutes, a roughly $20\times$ speedup. These MCQA results highlight PLOT's utility for accurate gradient-free handles and coarse localization that makes learned DAS substantially cheaper.


\begin{table}[!t]
\centering
\scriptsize
\setlength{\tabcolsep}{1.8pt}
\renewcommand{\arraystretch}{1.06}
\caption{MCQA results for IIA and runtime over five seeds. Layer/dim columns report seed~0, where dimension denotes calibrated PLOT support, DBM feature count, or DAS subspace dimension.}
\vspace{-3mm}
\label{tab:mcqa-results}

\resizebox{\linewidth}{!}{%
\begin{tabular}{lcccccc}
\toprule
Method & AP & AT & Avg. & Runtime (min) & AP layer/dim & AT layer/dim \\
\midrule

\onePLOTnat{}
& $0.4290 \pm 0.3381$
& $0.7620 \pm 0.2958$
& $0.5955 \pm 0.2779$
& $112.85 \pm 2.64$
& L5/576
& L20/1152 \\

\onePLOTpca{}
& $0.7640 \pm 0.1130$
& $0.9120 \pm 0.0561$
& $0.8380 \pm 0.0435$
& $124.71 \pm 3.09$
& L14,17,20,21,23/705
& L20,21,24/846 \\

\twoPLOTnat{}
& $0.7630 \pm 0.1026$
& $0.9030 \pm 0.0231$
& $0.8330 \pm 0.0523$
& $49.11 \pm 3.93$
& L17/512
& L24/1024 \\

\twoPLOTpca{}
& $0.7640 \pm 0.1099$
& $\mathbf{0.9580 \pm 0.0115}$
& $0.8610 \pm 0.0526$
& $67.50 \pm 4.34$
& L17/141
& L24/282 \\

\midrule

DBM canonical
& $0.7880 \pm 0.0582$
& $0.6610 \pm 0.0096$
& $0.7245 \pm 0.0264$
& $13.63 \pm 0.45$
& L17/1152
& L23/1142 \\

DBM PCA
& $0.7700 \pm 0.0581$
& $0.9440 \pm 0.0129$
& $0.8570 \pm 0.0325$
& $15.12 \pm 0.36$
& L17/219
& L23/246 \\

DBM SAE
& $0.7130 \pm 0.0512$
& $0.7990 \pm 0.0323$
& $0.7560 \pm 0.0142$
& $14.96 \pm 0.77$
& L20/109
& L23/166 \\

\midrule

PLOT-DAS
& $0.8810 \pm 0.0542$
& $0.8690 \pm 0.0500$
& $0.8750 \pm 0.0286$
& $25.27 \pm 1.60$
& L17/256
& L24/128 \\

Full DAS
& $\mathbf{0.9220 \pm 0.0282}$
& $0.8700 \pm 0.0296$
& $\mathbf{0.8960 \pm 0.0213}$
& $509.60 \pm 23.88$
& L17/32
& L20/512 \\

\bottomrule
\end{tabular}
}
\vspace{-3mm}
\end{table}

\section{Discussion and Concluding Remarks}
\label{sec:discussion}

We proposed PLOT, a gradient-free approach for joint correspondence discovery via OT matching of intervention effects. The coupling can be calibrated into executable intervention handles or used progressively to restrict subsequent localization and guide subspace learning. Our experiments show that progressive localization can improve accuracy under handle-dimension limits and reduce computation, while PLOT-guided DAS approaches full-DAS accuracy at substantially lower cost.

Future directions include sharper localization rules that exploit finer resolutions and hierarchical signature constructions. Scaling PLOT to larger models will also require engineering around batching, activation caching, approximate signatures, and distributed transport computation \citep{wang2023neuralegw,tsur2025neuralestimationscalingentropic}. Unlike subspace methods such as DAS, PLOT's matching formulation need not operate directly on the neural network and can instead align a causal model with an interpretable feature representation thereof. A natural direction is to apply PLOT to features in Anthropic's cross-layer transcoder replacement models or even to nodes in their attribution graphs \citep{ameisen2025circuit}, with the goal of automating attribution graph discovery or compression.

\bibliography{references}
\bibliographystyle{plainnat}


\appendix
\crefalias{section}{appendix}
\crefalias{subsection}{appendix}

\newpage\noindent\LARGE Appendix
\normalsize

\section{Experimental Details}
\label{app:tasks}
\subsection{Evaluation Conventions}\label{appen:test}

After calibrating $K$ and $\lambda$ for each abstract variable $Z_i$, we evaluate on separate holdout test sets $\mathcal{D}_\mathrm{te}$:
\begin{itemize}[leftmargin=1.3em,itemsep=0.7pt]
\vspace{-2mm}
    \item \textbf{\boldmath$Z_i$-sensitivity set:} containing pairs $(x^b,x^s)$ for which $x^s$ changes the value of $Z_i$ relative to $x^b$.
    \item \textbf{\boldmath$Z_i$-invariance set:} containing pairs $(x^b,x^s)$ for which $x^s$ leaves the value of $Z_i$ fixed.
\end{itemize}
We evaluate each calibrated handle separately on both sets, checking both sensitivity to interventions that change $Z_i$ and invariance to interventions that leave $Z_i$ unchanged. This differs from prior works \citep{geiger2024das, wu2024boundlessdas, mueller2025mib} which mixed sensitive and invariant pairs, leading to test accuracies that are less interpretable. Indeed, a method that learns no handle can score perfectly on invariance-heavy test sets, yielding high average accuracy while being weak on sensitivity. Unless stated otherwise, main-text accuracies average over all $2m$ sensitive and invariant test sets (two for each variable $Z_1,\ldots,Z_m$).



\subsection{Runtime Conventions}
We report serial method runtime per seed, starting after loading or training the factual model and obtaining data banks. Runtime includes method-specific preparation (such as effect signatures and PCA), coupling computation or DAS training, calibration over the reported search grid, and evaluation of the selected handle on the held-out test bank. For certain multi-stage methods, we may reuse handles and runtimes obtained from shared initial stages. Reported method runtime therefore underestimates the full wall-clock time of the experiment.

\subsection{Hierarchical Equality (HEQ)}
\label{app:tasks-heq}

\paragraph{Neural backbone training.}

We encode each tuple $(W,X,Y,Z)\in[100]^4$ using a fixed embedding map $e:[100]\to\R^4$ and concatenate the embeddings into $x=[e(W)\ e(X)\ e(Y)\ e(Z)]\in\R^{16}$. The neural model is a ReLU MLP with three hidden layers of width $16$, resulting in $850$ trainable parameters. We train the MLP with cross-entropy loss using Adam \citep{kingma2017adammethodstochasticoptimization}, learning rate $10^{-3}$, and batch size $1024$ for three epochs on $1{,}048{,}576$ factual examples, matching the setup in \citet{geiger2024das}. The model reaches exact accuracy $1.0$ on a disjoint validation set of size $10{,}000$.

\paragraph{Pair bank construction.}
We construct the HEQ fit, calibration, and test banks as follows:
\begin{itemize}[leftmargin=1.3em,itemsep=0.7pt]
\vspace{-2mm}
\item \textbf{Fit bank:} $1000$ pairs from a mixed policy with roughly 50\% pairs sensitive in either $z_{WX}$ or $z_{YZ}$, and 50\% invariant in both.
\item \textbf{Calibration bank:} $1000$ pairs, half only $z_{WX}$-sensitive and half only $z_{YZ}$-sensitive.
\item \textbf{Test banks:} four holdout banks of size $1000$, one sensitive and one invariant for each of $z_{WX}$ and~$z_{YZ}$.
\end{itemize}

\paragraph{Effect signature construction.}
We let $d^{\mathrm{abs}}(y,y')=y-y'$ where outputs are one-hot over two classes, and $d^{\mathrm{nn}}(y,y')=\operatorname{softmax}(y)-\operatorname{softmax}(y')$ where outputs are logits over two classes. Thus we compute $\Delta_{i,t}^{\mathrm{abs}},\Delta_{j,t}^{\mathrm{nn}}\in \R^2$ for each of the $T_{\mathrm{ft}}=1000$ fit pairs. Next, we take $\operatorname{Agg}$ as simply concatenation of the effects in a common fit-pair order without averaging or normalization, giving $u_i,v_j\in\mathbb{R}^{2000}$.

\paragraph{Hardware and runtime details.}
We run all methods locally on an Apple M1 Pro with 8 CPU cores and 16GB of memory. Runtime excludes factual-model training and includes the full parameter searches:

\begin{itemize}[leftmargin=1.3em,itemsep=0.7pt]
\vspace{-2mm}
\item \textbf{PLOT:} runtime includes effect signature construction over $\mathcal{D}_{\mathrm{ft}}$ for the two abstract variables and $3\times16=48$ neural sites, Sinkhorn fitting of an EOT coupling for each $\varepsilon\in\{0.01,0.25,2,4,8,16\}$, calibration of each coupling row over $K\in\{1,\ldots,16\}$ and $\lambda\in\{1,\ldots,16\}$ on $\mathcal{D}_{\mathrm{cal}}$, and final evaluation on $\mathcal{D}_{\mathrm{te}}$ of the calibrated intervention. This gives $2\times6\times16\times16=3072$ calibration settings per seed across the two variables.
\item \textbf{DAS:} for learning rate $10^{-3}$ and maximum epoch budget $N=1000$ with early stopping, runtime includes training rotations on $\mathcal{D}_{\mathrm{ft}}$ for each abstract variable, hidden layer $\ell\in\{1,2,3\}$, and subspace dimension $r\in\{1,\ldots,16\}$, calibration over all $(\ell,r)$ pairs on $\mathcal{D}_{\mathrm{cal}}$, and final evaluation of the selected rotation on $\mathcal{D}_{\mathrm{te}}$.
\end{itemize}
The accuracy and runtime in \Cref{fig:heq-joint-results} report means and standard deviations across seeds~1--10.

\paragraph{Heatmap figure details.}
To generate \Cref{fig:heq-handle-heatmaps}, we select one HEQ seed (namely, seed~7) to display learned handles. The accuracies and serial runtimes for this seed are displayed in \Cref{tab:heq-joint-results-seed7}.

The PLOT panels show matching weights over the $3\times16$ hidden neurons of the HEQ MLP. For PLOT before calibration, we show the coupling rows for $z_{WX}$ and $z_{YZ}$ over the $48$ candidate sites, renormalized to $[0,1]$ for visualization. For PLOT after calibration, we display only the weights on the calibrated top-$K$ support, again renormalized to $[0,1]$. Calibration selects the shared entropic parameter $\varepsilon=2$, with $(K,\lambda)=(6,6)$ for $z_{WX}$ and $(5,5)$ for $z_{YZ}$.

For DAS, we display the best calibration-selected rotation after sweeping layers $\ell$ and subspace dimensions $r$. For $z_{WX}$, $\ell=1$ and $r=10$, while for $z_{YZ}$, $\ell=1$ and $r=4$. In the canonical basis panel, we visualize each neuron's contribution to the selected intervention subspace, normalized to $[0,1]$. In the rotated basis panel, we mark the first $r$ rotated coordinates in layer $\ell$, which are the actual coordinates intervened on by DAS.

\begin{table}[!t]
    \centering
    \small
    \caption{HEQ results for seed~7, used to generate \Cref{fig:heq-handle-heatmaps}. Serial runtime includes the full searches, calibration, and test evaluation, but excludes factual model training and model/bank loading.}
    \label{tab:heq-joint-results-seed7}
    \begin{tabular*}{\textwidth}{@{\extracolsep{\fill}}lcccc>{\raggedright\arraybackslash}p{4cm}@{}}
        \toprule
        Method & $z_{WX}$ & $z_{YZ}$ & Average & Serial runtime (s) & Best setting \\
        \midrule
        \PLOT{}
        & $0.9980$
        & $0.9930$
        & $0.9955$
        & $3.80$
        & $\varepsilon=2$, $K=(6,5)$, $\lambda=(6,5)$ \\
        DAS
        & $0.9960$
        & $0.9900$
        & $0.9930$
        & $184.04$
        & Layer-1 sites with subspace dimensions $10$ for $z_{WX}$ and $4$ for $z_{YZ}$ \\
        \bottomrule
    \end{tabular*}
\end{table}

\begin{wrapfigure}{r}{0.43\linewidth}
    \vspace{-4.75mm}
    \centering
    \includegraphics[width=.99\linewidth]{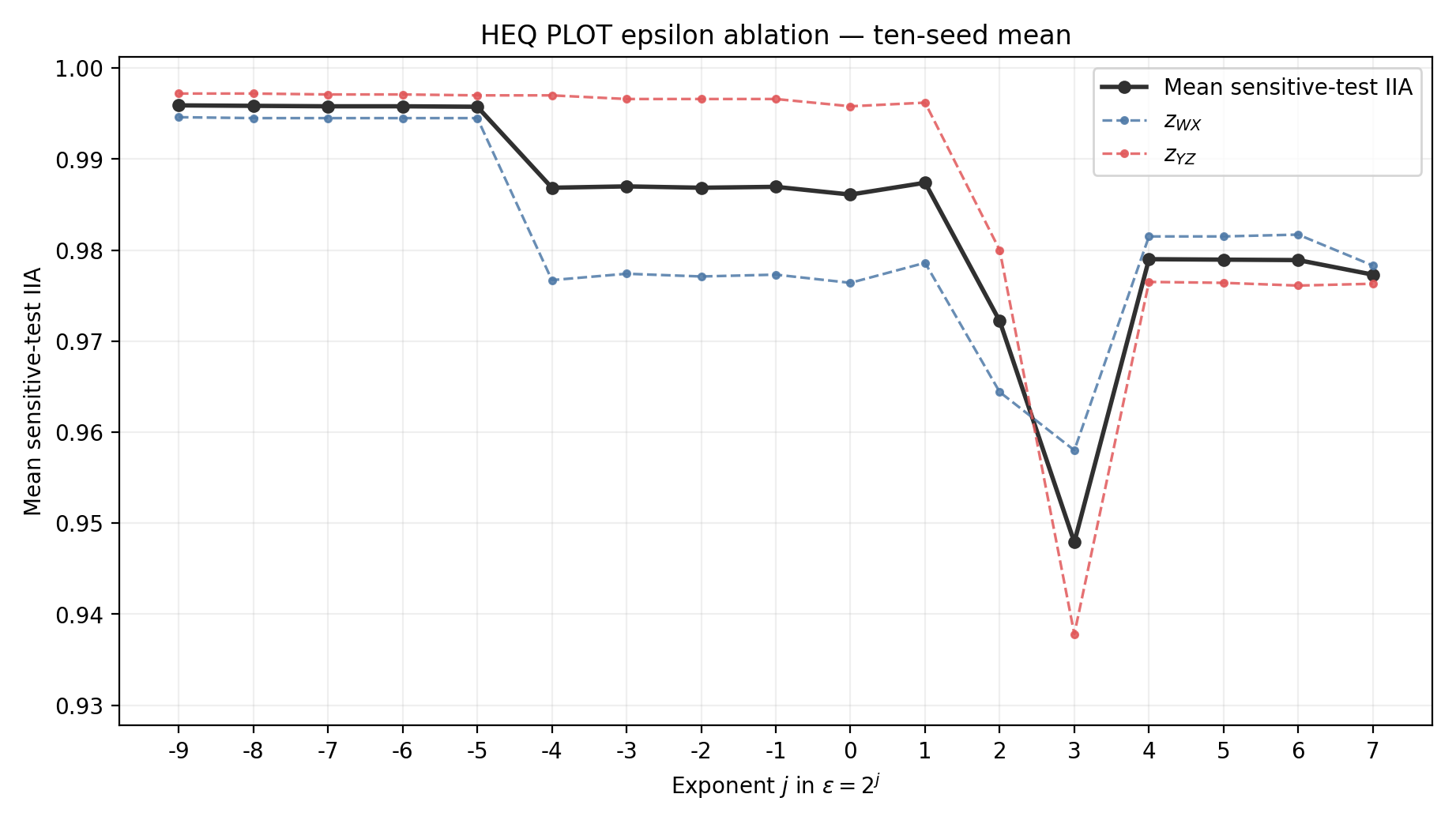}
    \caption{Sensitivity to the entropic parameter $\varepsilon$, averaged over ten seeds.}
\label{fig:heq-ot-epsilon}
\end{wrapfigure}

\paragraph{Effect of entropic regularization.}
We test sensitivity to the entropic regularization parameter $\varepsilon$ in \eqref{eq:eot} by rerunning the HEQ PLOT pipeline on the same backbone and calibration protocol while varying only $\varepsilon$. At each fixed $\varepsilon$, we select $K$ and $\lambda$ on $\mathcal{D}_{\mathrm{cal}}$ and evaluate the resulting handles on the held-out test banks. As shown in \Cref{fig:heq-ot-epsilon}, smaller values of $\varepsilon$ perform best, with mean sensitive-test IIA remaining near $0.996$ for $\varepsilon=2^j$, $j=-9,\ldots,-5$. Performance decreases as the regularization grows, with the largest drop around $\varepsilon=2^3$, before partially recovering for larger values.



\subsection{Binary Addition}
\label{app:tasks-addition}

\paragraph{Neural backbone training.}
We use a single-layer GRUCell recurrent network, which utilizes one gated recurrent unit (GRU) cell\footnote{\url{https://docs.pytorch.org/docs/2.11/generated/torch.nn.GRUCell.html}}. The input consists of two $4$-bit numbers
\[
a=(a_0,a_1,a_2,a_3),\qquad b=(b_0,b_1,b_2,b_3),
\]
processed from least to most significant bit. At timestep $\ell$,
\[
x_\ell=[a_\ell,b_\ell]\in\mathbb{R}^2,\qquad
h_\ell=\mathrm{GRUCell}(x_\ell,h_{\ell-1})\in\mathbb{R}^d,\qquad h_{-1}=0.
\]
The hidden state therefore carries information from lower-order bits to higher-order bits, enabling carry propagation. Sum-bit logits are read from each timestep,
\[
y_\ell^{(S)}=w_S^\intercal h_\ell+b_S,\qquad \ell=0,1,2,3,
\]
and the final carry logit is read from the last state,
\[
y^{(C)}=w_C^\intercal h_3+b_C,\qquad
y=[y^{(C)},y_3^{(S)},y_2^{(S)},y_1^{(S)},y_0^{(S)}]\in\mathbb{R}^5.
\]
We train a model with $d = 16$ using binary cross entropy and Adam at learning rate $2\times 10^{-2}$, batch size $64$, and $250$ epochs with early stopping. The model achieves exact accuracy on all $2^4\times2^4=256$ addition pairs.


\paragraph{Pair bank construction.} There are $2^4\times 2^4=256$ possible inputs, which we collect as a bank of base examples. For each base example, we use a policy to sample $26$ source examples: 
\begin{itemize}[leftmargin=1.3em,itemsep=0.7pt]
\vspace{-2mm}
\item $8$ source examples where each one flips the value of one of the $8$ input bits, $a_0,\ldots,a_3,b_0,\ldots,b_3$.
\item $18$ carry-targeted source examples: $3$ that change the value of $C_1$, $5$ for $C_2$, $7$ for $C_3$, and $3$ for $C_4$. 
\end{itemize}
We use a $128-64-64$ split for the $256$ base examples, which means $128\times 26 = 3328$ pairs for the fit set, $64\times 26=1664$ for the calibration set, and $64\times 26 = 1664$ for the test set. 

For the test set, we record for each pair whether or not it changed the value of $C_1$, $C_2$, and $C_3$, the three abstract variables of interest. Then, we partition the $1664$ test pairs into sensitive and invariant sets for each of these variables.

\paragraph{Effect signature construction.}
We let $d^{\mathrm{abs}}(y,y')=y-y'$ where outputs are five-bit binary numbers, and $d^{\mathrm{nn}}(y,y')=\sigma(y)-\sigma(y')$ where outputs are logits for each of the 
five bits. Thus we compute $\Delta_{i,t}^{\mathrm{abs}},\Delta_{j,t}^{\mathrm{nn}}\in \R^5$ for each of the $T_\mathrm{ft} = 3328$ fit pairs. The fit bank contains $26$ source-construction families with $128$ pairs per family, so we take $\operatorname{Agg}$ to be the mean five-dimensional effect within each family, concatenate over the $26$ means, and $\ell_2$-normalize the resulting vector. Thus $u_i,v_j\in\mathbb{R}^{130}$ for full-timestep, native-block, and PCA-band sites.

\paragraph{Hardware, runtime, and algorithm details.} We run all methods locally on an Apple M1 Pro with 8 CPU cores and 16GB of memory. The algorithm and runtime for each method are detailed below.

\begin{algorithm}[htbp]
\caption{Binary Addition: \onePLOTnat{}}
\label{alg:addition-one-plot-nat}
\begin{algorithmic}[1]
\Require Splits $\mathcal{D}_{\mathrm{ft}}$, $\mathcal{D}_{\mathrm{cal}}$, and $\mathcal{D}_{\mathrm{te}}$, factual pre-trained GRUCell with hidden size $d$
\State Start timer
\State Construct effect signatures over $\mathcal{D}_{\mathrm{ft}}$ for abstract carry variables $C_1,C_2,C_3$
\For{resolution $r \in \{1,2,4,\ldots,d\}$}
    \State Partition $h_0,\ldots,h_3$ into neural sites with resolution $r$
    \State Construct effect signatures over $\mathcal{D}_{\mathrm{ft}}$ for the partitioned neural sites
    \For{$\varepsilon\in\{0.003, 0.01, 0.03, 0.1\}$}
        \State Compute the Sinkhorn EOT coupling in $\R^{3 \times 4\lfloor d/r \rfloor}$ using $\varepsilon$
        \For{carry variable $C_i \in \{C_1,C_2,C_3\}$}
            \For{$(\text{top-}K,\lambda) \in \{1,2,4\} \times \{0.25, 0.5, 1,2,4,8\}$}
                \State Evaluate the candidate intervention handle $H_i^{K,\lambda}$ for $C_i$ on $\mathcal{D}_{\mathrm{cal}}$
            \EndFor
        \EndFor
    \EndFor
\EndFor
\State Evaluate the best calibrated handles $H_1^\star, \ldots, H_3^\star$ on $\mathcal{D}_{\mathrm{te}}$
\State End timer and report \onePLOTnat{} runtime
\end{algorithmic}
\end{algorithm}

\begin{algorithm}[htbp]
\caption{Binary Addition: \onePLOTpca{}}
\label{alg:addition-one-plot-pca}
\begin{algorithmic}[1]
\Require Splits $\mathcal{D}_{\mathrm{ft}}$, $\mathcal{D}_{\mathrm{cal}}$, and $\mathcal{D}_{\mathrm{te}}$, factual pre-trained GRUCell with hidden size $d$
\State Start timer
\State Construct effect signatures over $\mathcal{D}_{\mathrm{ft}}$ for abstract carry variables $C_1,C_2,C_3$
\State Collect hidden states from $\mathcal{D}_{\mathrm{ft}}$ for each timestep $h_0,\ldots,h_3$
\State Compute the centered PCA rotations with full rank; store them as $R_0, \ldots, R_3$
\For{band count $b \in \{1,2,4,\ldots,d\}$}
    \State Partition the $d$ PCA directions for each timestep into $b$ disjoint neural sites
    \State Set $W_j$ to be the columns of $R_{h_j}$ corresponding to $s_j$, where $h_j$ is the timestep containing $s_j$ 
    \State Construct effect signatures over $\mathcal{D}_{\mathrm{ft}}$ for the PCA-band neural sites
    \For{$\varepsilon \in \{0.003, 0.01, 0.03, 0.1\}$}
    \State Compute the Sinkhorn EOT coupling in $\R^{3 \times 4b}$ using $\varepsilon$
    \For{carry variable $C_i \in \{C_1, C_2, C_3\}$}
        \For{$(\text{top-}K,\lambda) \in \{1,2,4\} \times \{0.25, 0.5, 1,2,4,8\}$}
            \State Evaluate the candidate intervention handle $H_i^{K,\lambda}$ for $C_i$ on $\mathcal{D}_{\mathrm{cal}}$
        \EndFor
    \EndFor
    \EndFor
\EndFor
\State Evaluate the best calibrated handles $H_1^\star, \ldots, H_3^\star$ on $\mathcal{D}_{\mathrm{te}}$
\State End timer and report \onePLOTpca{} runtime
\end{algorithmic}
\end{algorithm}

\begin{algorithm}[htbp]
\caption{Binary Addition: \twoPLOTnat{}}
\label{alg:addition-two-plot-nat}
\begin{algorithmic}[1]
\Require Splits $\mathcal{D}_{\mathrm{ft}}$, $\mathcal{D}_{\mathrm{cal}}$, and $\mathcal{D}_{\mathrm{te}}$, factual pre-trained GRUCell with hidden size $d$
\State Start timer
\State \underline{\textbf{Stage 1:}}
\State Construct effect signatures over $\mathcal{D}_{\mathrm{ft}}$ for the four neural timestep activations $h_0,\ldots,h_3$
\State Construct effect signatures over $\mathcal{D}_{\mathrm{ft}}$ for abstract carry variables $C_1,C_2,C_3$
\For{$\varepsilon \in \{0.01, 0.03\}$}
\State Compute the Sinkhorn EOT coupling in $\R^{3 \times 4}$ using $\varepsilon$
\For{carry variable $C_i \in \{C_1,C_2,C_3\}$}
    \State Extract the top timestep $\ell_i$ for row $C_i$
    \State Evaluate timestep $h_{\ell_i}$ for $C_i$ on $\mathcal{D}_{\mathrm{cal}}$
\EndFor
\EndFor
\State Choose $\{\ell_1,\ell_2,\ell_3\}$ that achieves highest mean calibration score across choices of $\varepsilon$
\State \underline{\textbf{Stage 2:}}
\For{$\ell \in \{\ell_1,\ell_2,\ell_3\}$}
\For{resolution $r \in \{1,2,4,\ldots,d\}$}
    \State Partition $h_\ell$ into neural sites with resolution $r$
    \State Construct effect signatures over $\mathcal{D}_{\mathrm{ft}}$ for the partitioned neural sites
    \For{$\varepsilon \in \{0.003, 0.01, 0.03, 0.1\}$}
        \State Compute the Sinkhorn EOT coupling in $\R^{3 \times \lfloor d/r \rfloor}$ using $\varepsilon$
        \For{$C_i$ with top timestep $\ell$}
            \For{$(\text{top-}K,\lambda) \in \{1,2,4\} \times \{0.25, 0.5, 1,2,4,8\}$}
                \State Evaluate the candidate intervention handle $H_i^{K,\lambda}$ for $C_i$ on $\mathcal{D}_{\mathrm{cal}}$
            \EndFor
        \EndFor
    \EndFor
    \EndFor
\EndFor
\State Evaluate the best calibrated handles $H_1^\star, \ldots, H_3^\star$ on $\mathcal{D}_{\mathrm{te}}$
\State End timer and report \twoPLOTnat{} runtime
\end{algorithmic}
\end{algorithm}

\begin{algorithm}[htbp]
\caption{Binary Addition \twoPLOTpca{}}
\label{alg:addition-two-plot-pca}
\begin{algorithmic}[1]
\Require Splits $\mathcal{D}_{\mathrm{ft}}$, $\mathcal{D}_{\mathrm{cal}}$, and $\mathcal{D}_{\mathrm{te}}$, factual pre-trained GRUCell with hidden size $d$
\State Start timer
\State \underline{\textbf{Stage 1:}}
\State Construct effect signatures over $\mathcal{D}_{\mathrm{ft}}$ for the four neural timestep activations $h_0,\ldots,h_3$
\State Construct effect signatures over $\mathcal{D}_{\mathrm{ft}}$ for abstract carry variables $C_1,C_2,C_3$
\For{$\varepsilon \in \{0.01, 0.03\}$}
\State Compute the Sinkhorn EOT coupling in $\R^{3 \times 4}$ using $\varepsilon$
\For{carry variable $C_i \in \{C_1,C_2,C_3\}$}
    \State Extract the top timestep $\ell_i$ for row $C_i$
    \State Evaluate timestep $h_{\ell_i}$ for $C_i$ on $\mathcal{D}_{\mathrm{cal}}$
    \EndFor
\EndFor
\State Choose $\{\ell_1,\ell_2,\ell_3\}$ that achieves the highest mean calibration score across choices of $\varepsilon$
\State \underline{\textbf{Stage 2:}}
\For{$\ell \in \{\ell_1,\ell_2,\ell_3\}$}
    \State Collect hidden states from $\mathcal{D}_{\mathrm{ft}}$ at timestep $h_\ell$
    \State Compute the centered PCA rotation with full rank; store it as $R_\ell$
    \For{band count $b \in \{1,2,4,\ldots,d\}$}
        \State Partition the $d$ PCA directions at $h_\ell$ into $b$ disjoint neural sites
        \State For each $j$, set $W_j$ to be the columns of $R_\ell$ corresponding to the PCA-band $s_j$
        \State Construct effect signatures over $\mathcal{D}_{\mathrm{ft}}$ for the PCA-band neural sites
        \For{$\varepsilon \in \{0.003, 0.01, 0.03, 0.1\}$}
        \State Compute the Sinkhorn EOT coupling in $\R^{3 \times b}$ using $\varepsilon$
        \For{$C_i$ with top timestep $\ell$}
            \For{$(\text{top-}K,\lambda) \in \{1,2,4\} \times \{0.25, 0.5, 1,2,4,8\}$}
                \State Evaluate the candidate intervention handle $H_i^{K,\lambda}$ for $C_i$ on $\mathcal{D}_{\mathrm{cal}}$
            \EndFor
        \EndFor
        \EndFor
    \EndFor
\EndFor
\State Evaluate the best calibrated handles $H_1^\star, \ldots, H_3^\star$ on $\mathcal{D}_{\mathrm{te}}$
\State End timer and report \twoPLOTpca{} runtime
\end{algorithmic}
\end{algorithm}

\begin{algorithm}[htbp]
\caption{Binary Addition \PLOTdas{}}
\label{alg:addition-plot-das}
\begin{algorithmic}[1]
\Require DAS learning rate $\mathrm{lr}=10^{-2}$, DAS maximum epochs $N=100$
\Require Splits $\mathcal{D}_{\mathrm{ft}}$, $\mathcal{D}_{\mathrm{cal}}$, and $\mathcal{D}_{\mathrm{te}}$, factual pre-trained GRUCell with hidden size $d$
\State Start timer
\State \underline{\textbf{Stage 1:}}
\State Construct effect signatures over $\mathcal{D}_{\mathrm{ft}}$ for the four neural timestep activations $h_0,\ldots,h_3$
\State Construct effect signatures over $\mathcal{D}_{\mathrm{ft}}$ for abstract carry variables $C_1,C_2,C_3$
\For{$\varepsilon \in \{0.01, 0.03\}$}
\State Compute the Sinkhorn EOT coupling in $\R^{3 \times 4}$ using $\varepsilon$
\For{carry variable $C_i \in \{C_1,C_2,C_3\}$}
    \State Extract the top timestep $\ell_i$ for row $C_i$
    \State Evaluate timestep $h_{\ell_i}$ for $C_i$ on $\mathcal{D}_{\mathrm{cal}}$
    \EndFor
\EndFor
\State Choose $\{\ell_1,\ell_2,\ell_3\}$ that achieves the highest mean calibration score across choices of $\varepsilon$
\State \underline{\textbf{Stage 2:}}
\For{$C_i \in \{C_1,C_2,C_3\}$}
    \For{$k \in \{1,2,4,\ldots,d\}$}
        \State Train a DAS rotation for $C_i$ on $\mathcal{D}_{\mathrm{ft}}$ at timestep $h_{\ell_i}$ with subspace intervention size $k$
        \State Evaluate the rotation handle for $C_i$ on $\mathcal{D}_{\mathrm{cal}}$
    \EndFor
\EndFor
\State Evaluate the best calibrated rotations for $C_1,C_2,C_3$ on $\mathcal{D}_{\mathrm{te}}$
\State End timer and report \PLOTdas{} runtime
\end{algorithmic}
\end{algorithm}

\begin{algorithm}[htbp]
\caption{Binary Addition Full DAS}
\label{alg:addition-full-das}
\begin{algorithmic}[1]
\Require DAS learning rate $\mathrm{lr}=10^{-2}$, DAS maximum epochs $N=100$
\Require Splits $\mathcal{D}_{\mathrm{ft}}$, $\mathcal{D}_{\mathrm{cal}}$, and $\mathcal{D}_{\mathrm{te}}$, factual pre-trained GRUCell with hidden size $d$
\State Start timer
\For{$C_i \in \{C_1,C_2,C_3\}$}
    \For{timestep $\ell \in \{0,1,2,3\}$}
        \For{$k \in \{1,2,4,\ldots, d\}$}
            \State Train a DAS rotation for $C_i$ on $\mathcal{D}_{\mathrm{ft}}$ at timestep $h_\ell$ with subspace intervention size $k$
            \State Evaluate the rotation handle for $C_i$ on $\mathcal{D}_{\mathrm{cal}}$
        \EndFor
    \EndFor
\EndFor
\State Evaluate the best calibrated rotations for $C_1,C_2,C_3$ on $\mathcal{D}_{\mathrm{te}}$
\State End timer and report Full DAS runtime
\end{algorithmic}
\end{algorithm}

\newpage

\subsection{Multiple-Choice Question Answering (MCQA)}
\label{appen:mcqa}

\paragraph{Pair bank construction.} 

For the MCQA task, MIB-Bench \citep{mueller2025mib} open-sources a dataset\footnote{\url{https://huggingface.co/datasets/mib-bench/copycolors_mcqa}} used to train and test their methods. However, this public dataset only contains $210$ rows, so we synthetically generate a dataset\footnote{
\url{https://huggingface.co/datasets/jchang153/copycolors_mcqa}
} with $10{,}000$ rows in the same format as their dataset. See \Cref{tab:mcqa-selected-counterfactuals} for a base example and associated counterfactuals. Recall that the causal model in Figure~\ref{fig:mcqa-causal-model} contains two abstract variables: $\mathrm{AP}$ and $\mathrm{AT}$. As in MIB-Bench, we consider the following three counterfactuals:

\begin{itemize}[leftmargin=1.3em,itemsep=0.7pt]
\vspace{-2mm}
\item \texttt{answerPosition}: permute the positions of the answer choice colors $\mathrm{Col_i}$, without changing the answer choice symbols $\mathrm{Sym_i}$. This typically affects both $\mathrm{AP}$ and $\mathrm{AT}$.
\item \texttt{randomLetter}: randomly sample four different letters of the alphabet to replace the answer choice symbols $\mathrm{Sym_i}$. This typically affects $\mathrm{AT}$.
\item \texttt{answerPosition + randomLetter}: apply both counterfactuals above at the same time. This typically affects both $\mathrm{AP}$ and $\mathrm{AT}$.
\end{itemize}

Before constructing pair banks, we filter examples to ensure that Gemma-2-2B correctly completes the prompts before any interventions. We first sample $2{,}000$ rows from our dataset, forming $2{,}000 \times 3 = 6{,}000$ (base, source) pairs from the three counterfactuals. Then we pass each pair into Gemma-2-2B: a (base, source) pair passes the filter if Gemma-2-2B correctly completes both the base and source examples, up to capitalization, whitespace, and punctuation. Finally, we sample the following fit, calibration, and test banks: 
\begin{itemize}[leftmargin=1.3em,itemsep=0.7pt]
\vspace{-2mm}
\item Fit bank: $200$ pairs sampled randomly from the filtered bank.
\item Calibration banks: $200$ sensitive pairs for each $\mathrm{AP}$ and $\mathrm{AT}$.
\item Test banks: $200$ sensitive pairs for each $\mathrm{AP}$ and $\mathrm{AT}$. 
\end{itemize}
We ensure that no two banks contain (base, source) pairs with the same base question.

\begin{table}[htbp]
\centering
\caption{An MCQA example and selected associated counterfactuals.}
\label{tab:mcqa-selected-counterfactuals}
\renewcommand{\arraystretch}{1.15}
\begin{tabular}{p{0.27\linewidth} p{0.42\linewidth} c}
\toprule 
\rule{0pt}{16pt} \textbf{Counterfactual} & \textbf{Text} & \textbf{Correct Completion} \\[6pt]
\midrule
Original Prompt &
\textit{Question: A banana is yellow. What color is a banana? A.\ red\textbackslash nB.\ blue\textbackslash nC.\ yellow\textbackslash nD.\ green\textbackslash nAnswer:} & C \\
\midrule
\texttt{answerPosition} &
\textit{Question: A banana is yellow. What color is a banana? A.\ green\textbackslash nB.\ yellow\textbackslash nC.\ red\textbackslash nD.\ blue\textbackslash nAnswer:} & B \\
\midrule
\texttt{randomLetter} &
\textit{Question: A banana is yellow. What color is a banana? H.\ red\textbackslash nT.\ blue\textbackslash nP.\ yellow\textbackslash nX.\ green\textbackslash nAnswer:} & P \\
\midrule
\texttt{answerPosition + randomLetter} &
\textit{Question: A banana is yellow. What color is a banana? H.\ green\textbackslash nT.\ yellow\textbackslash nP.\ red\textbackslash nX.\ blue\textbackslash nAnswer:} & T \\
\bottomrule
\end{tabular}
\end{table}

\paragraph{Effect signature construction.} See \Cref{app:ablations-signatures}, where we compare three variants of signature construction as an ablation experiment---family-wise, KL, and TV. The experiments in our main text utilize the family-wise signature. 

\paragraph{OT vs. UOT details.}
In \Cref{tab:mcqa-ot-uot-selected-layers}, we compare OT and UOT as Stage-A layer-selection methods for the final-token residual stream. Each method produces a $2\times26$ coupling, with rows corresponding to $\mathrm{AP}$ and $\mathrm{AT}$ and columns corresponding to transformer layers. For each row, we take the top-$6$ layers by mass, calibrate the associated full-layer interventions on that abstract variable, and report the layer with the highest calibration accuracy. Because standard OT must distribute mass across all 26 layer sites using only two abstract rows, it yields weak layer selection for $\mathrm{AP}$. In contrast, UOT can leave irrelevant layers unmatched, which allows it to select high-scoring $\mathrm{AP}$ layers across all five seeds.



\begin{table}[htbp]
\caption{Selected MCQA layers and accuracies for Stage A OT vs UOT across five seeds.}
\centering
\begin{tabular}{c cc cc}
\toprule
& \multicolumn{2}{c}{OT} & \multicolumn{2}{c}{UOT} \\
\cmidrule(lr){2-3}\cmidrule(lr){4-5}
Seed & $\mathrm{AP}$ & $\mathrm{AT}$ & $\mathrm{AP}$ & $\mathrm{AT}$ \\
\midrule
0 & L6 (0.000) & L24 (0.970) & L17 (0.765) & L24 (0.970) \\
1 & L6 (0.000) & L24 (0.975) & L17 (0.775) & L24 (0.975) \\
2 & L6 (0.000) & L24 (0.965) & L17 (0.770) & L24 (0.965) \\
3 & L6 (0.000) & L24 (0.975) & L19 (0.695) & L24 (0.975) \\
4 & L6 (0.000) & L24 (0.985) & L19 (0.685) & L24 (0.985) \\
\bottomrule
\end{tabular}
\label{tab:mcqa-ot-uot-selected-layers}
\end{table}

\paragraph{Intervention agreement relation.}
For $Z\in\{\mathrm{AP},\mathrm{AT}\}$, the abstract intervention output $y_{i,t}^{\mathrm{abs\_swap}}$ is a single uppercase answer letter for pair $(x_t^b,x_t^s)$. After applying a handle $H_i$ to the base prompt, let $L_{H_i,t}(v)$ denote the next-token logit assigned to token $v\in\mathcal V$. For MCQA, the task-dependent agreement relation $\sim$ in \eqref{EQ:IIA_cal} is
\[
y_{H_i}^{\mathrm{nn\_swap}}(x_t^b,x_t^s)
\sim
y_{i,t}^{\mathrm{abs\_swap}}
\quad\Longleftrightarrow\quad
\mathcal{N}\!\left(
\operatorname*{arg\,max}_{v\in\mathcal{V}} L_{H_i,t}(v)\right)
=
y_{i,t}^{\mathrm{abs\_swap}},
\]
where $\mathcal{V}$ is the full next-token vocabulary. The function $\mathcal{N}$ decodes the token then applies Unicode NFKC normalization, trims surrounding whitespace, and folds case to uppercase; it returns an invalid symbol unless the result is exactly one letter in A--Z. We use this agreement relation to compute the fraction of correct pairs in each target's $200$-pair test bank. The reported overall MCQA IIA is the arithmetic mean of the $\mathrm{AP}$ and $\mathrm{AT}$ test accuracies.

We note that this differs from MIB's evaluation procedure, which instead checks whether the entire decoded continuation at all contains the expected answer letter preceded by a space. This substring criterion does not require the continuation to be exactly the answer letter: it can count a longer output containing that letter as correct. We find this protocol too relaxed, and therefore report held-out IIA using exact agreement after normalizing the highest-logit token to a single letter. 

\paragraph{Hardware, runtime, and algorithm details.} We run all seeds and methods on NVIDIA A40 GPUs. The algorithm and runtime for each method are detailed below.

\begin{algorithm}[htbp]
\caption{MCQA: \onePLOTnat{}}
\label{alg:mcqa-one-plot-nat}
\begin{algorithmic}[1]
\Require Splits $\mathcal{D}_{\mathrm{ft}}$, $\mathcal{D}_{\mathrm{cal}}$, and $\mathcal{D}_{\mathrm{te}}$, factual pre-trained Gemma-2-2B
\State Start timer
\State Construct effect signatures over $\mathcal{D}_{\mathrm{ft}}$ for abstract variables $\mathrm{AP}, \mathrm{AT}$
\For{resolution $r \in \{32,64,128,256,576,1152\}$}
    \State Partition final-token residual layers $\ell_0,\ldots,\ell_{25}$ into neural sites with resolution $r$
    \State Construct effect signatures over $\mathcal{D}_{\mathrm{ft}}$ for the partitioned neural sites
    \For{$(\varepsilon,\beta_\mathrm{neural})\in\{0.5\}\times\{0.1,0.3,1,3\}$}
        \State Compute the Sinkhorn UOT coupling in $\R^{2 \times 26\lfloor 2304/r \rfloor}$ using $\varepsilon$ and $\beta_\mathrm{neural}$
        \For{$Z \in \{\mathrm{AP}, \mathrm{AT}\}$}
            \For{$(\text{top-}K,\lambda) \in \{1,2,3,4,5\} \times \{0.5, 1,2,4,8,12,20,30\}$}
                \If{handle size does not exceed 1152 neurons}
                \State Evaluate the candidate intervention handle $H^{K,\lambda}$ for $Z$ on $\mathcal{D}_{\mathrm{cal}}$
                \EndIf
            \EndFor
        \EndFor
    \EndFor
\EndFor
\State Evaluate the best calibrated handles $H_\mathrm{AP}^\star,  H_\mathrm{AT}^\star$ on $\mathcal{D}_{\mathrm{te}}$
\State End timer and report \onePLOTnat{} runtime
\end{algorithmic}
\end{algorithm}

\begin{algorithm}[htbp]
\caption{MCQA: \onePLOTpca{}}
\label{alg:mcqa-one-plot-pca}
\begin{algorithmic}[1]
\Require Splits $\mathcal{D}_{\mathrm{ft}}$, $\mathcal{D}_{\mathrm{cal}}$, and $\mathcal{D}_{\mathrm{te}}$, factual Gemma-2-2B
\State Start timer
\State Construct effect signatures over $\mathcal{D}_{\mathrm{ft}}$ for abstract variables $\mathrm{AP}, \mathrm{AT}$
\State Collect hidden states from $\mathcal{D}_{\mathrm{ft}}$ for each final-token residual layer $\ell_0,\ldots,\ell_{25}$
\State Compute the centered PCA rotations with retained empirical rank; store them as $R_0, \ldots, R_{25}$
\For{band count $b \in \{1,2,4,8,16,32,64\}$}
    \State Partition each layer's retained PCA directions into $b$ disjoint neural sites
    \State Set $W_j$ to be the columns of $R_{\ell_j}$ corresponding to $s_j$, where $\ell_j$ is the layer containing $s_j$ 
    \State Construct effect signatures over $\mathcal{D}_{\mathrm{ft}}$ for the PCA-band neural sites
    \For{$(\varepsilon, \beta_\mathrm{neural}) \in \{0.5\} \times \{0.1,0.3,1,3\}$}
    \State Compute the Sinkhorn UOT coupling in $\R^{2 \times 26b}$ using $\varepsilon$ and $\beta_\mathrm{neural}$
    \For{$Z \in \{\mathrm{AP}, \mathrm{AT}\}$}
        \For{$(\text{top-}K,\lambda) \in \{1,2,3,4,5\} \times \{0.5,1,2,4,8,12,20,30\}$}
            \If{handle size does not exceed 1152 PCA directions}
            \State Evaluate the candidate intervention handle $H^{K,\lambda}$ for $Z$ on $\mathcal{D}_{\mathrm{cal}}$
            \EndIf
        \EndFor
    \EndFor
    \EndFor
\EndFor
\State Evaluate the best calibrated handles $H_\mathrm{AP}^\star, H_\mathrm{AT}^\star$ on $\mathcal{D}_{\mathrm{te}}$
\State End timer and report \onePLOTpca{} runtime
\end{algorithmic}
\end{algorithm}

\begin{algorithm}[htbp]
\caption{MCQA: \twoPLOTnat{}}
\label{alg:mcqa-two-plot-nat}
\begin{algorithmic}[1]
\Require Splits $\mathcal{D}_{\mathrm{ft}}$, $\mathcal{D}_{\mathrm{cal}}$, and $\mathcal{D}_{\mathrm{te}}$, factual pre-trained Gemma-2-2B
\State Start timer
\State \underline{\textbf{Stage 1:}}
\State Construct effect signatures over $\mathcal{D}_{\mathrm{ft}}$ for final-token residual layers $\ell_0,\ldots,\ell_{25}$
\State Construct effect signatures over $\mathcal{D}_{\mathrm{ft}}$ for abstract variables $\mathrm{AP},\mathrm{AT}$
\For{$(\varepsilon,\beta_{\mathrm{neural}})\in\{0.5\}\times\{0.1,0.3,1,3\}$}
\State Compute the Sinkhorn UOT coupling in $\R^{2\times 26}$ using $\varepsilon$ and $\beta_{\mathrm{neural}}$
\For{$Z\in\{\mathrm{AP},\mathrm{AT}\}$}
\State Extract the six highest-mass layers for row $Z$
\State Evaluate their full-layer interventions for $Z$ on $\mathcal{D}_{\mathrm{cal}}$; retain highest-scoring layer $\ell_Z$
\EndFor
\EndFor
\State Choose $\{\ell_\mathrm{AP}, \ell_\mathrm{AT}\}$ that achieves highest mean calibration score across choices of $\varepsilon$ and $\beta_\mathrm{neural}$
\State \underline{\textbf{Stage 2:}}
\For{$\ell \in \{\ell_\mathrm{AP},\ell_\mathrm{AT}\}$}
\For{resolution $r\in\{32,64,128,256,576,1152\}$}
\State Partition $\ell$ into neural sites with resolution $r$
\State Construct effect signatures over $\mathcal{D}_{\mathrm{ft}}$ for the partitioned neural sites
\For{$(\varepsilon,\beta_{\mathrm{neural}})\in\{0.5\}\times\{0.1,0.3,1,3\}$}
\State Compute the Sinkhorn UOT coupling in $\R^{2\times\lfloor 2304/r\rfloor}$ using $\varepsilon$ and $\beta_{\mathrm{neural}}$
\For{$Z\in\{\mathrm{AP},\mathrm{AT}\}$ with top layer $\ell$}
\For{$(\text{top-}K,\lambda)\in\{1,2,3,4,5\}\times\{0.5,1,2,4,8,12,20,30\}$}
\If{handle size does not exceed 1152 neurons}
\State Evaluate the candidate intervention handle $H_Z^{K,\lambda}$ on $\mathcal{D}_{\mathrm{cal}}$
\EndIf
\EndFor
\EndFor
\EndFor
\EndFor
\EndFor
\State Evaluate the best calibrated handles $H_{\mathrm{AP}}^\star,H_{\mathrm{AT}}^\star$ on $\mathcal{D}_{\mathrm{te}}$
\State End timer and report \twoPLOTnat{} runtime
\end{algorithmic}
\end{algorithm}

\begin{algorithm}[htbp]
\caption{MCQA: \twoPLOTpca{}}
\label{alg:mcqa-two-plot-pca}
\begin{algorithmic}[1]
\Require Splits $\mathcal{D}_{\mathrm{ft}}$, $\mathcal{D}_{\mathrm{cal}}$, and $\mathcal{D}_{\mathrm{te}}$, factual pre-trained Gemma-2-2B
\State Start timer
\State \underline{\textbf{Stage 1:}}
\State Construct effect signatures over $\mathcal{D}_{\mathrm{ft}}$ for final-token residual layers $\ell_0,\ldots,\ell_{25}$
\State Construct effect signatures over $\mathcal{D}_{\mathrm{ft}}$ for abstract variables $\mathrm{AP},\mathrm{AT}$
\For{$(\varepsilon,\beta_{\mathrm{neural}})\in\{0.5\}\times\{0.1,0.3,1,3\}$}
\State Compute the Sinkhorn UOT coupling in $\R^{2\times 26}$ using $\varepsilon$ and $\beta_{\mathrm{neural}}$
\For{$Z\in\{\mathrm{AP},\mathrm{AT}\}$}
\State Extract the six highest-mass layers for row $Z$
\State Evaluate their full-layer interventions for $Z$ on $\mathcal{D}_{\mathrm{cal}}$; retain highest-scoring layer $\ell_Z$
\EndFor
\EndFor
\State Choose $\{\ell_\mathrm{AP}, \ell_\mathrm{AT}\}$ that achieves highest mean calibration score across choices of $\varepsilon$ and $\beta_\mathrm{neural}$
\State \underline{\textbf{Stage 2:}}
\For{$\ell \in \{\ell_\mathrm{AP}, \ell_\mathrm{AT}\}$}
\State Collect hidden states from $\mathcal{D}_{\mathrm{ft}}$ for final-token residual layer $\ell$
\State Compute the centered PCA basis with retained empirical rank; store it as $R_\ell$
\For{band count $b\in\{1,2,4,8,16,32,64\}$}
\State Partition the retained PCA directions at $\ell$ into $b$ disjoint neural sites
\State Set $W_j$ to be the columns of $R_{\ell}$ corresponding to $s_j$, for each PCA-band $s_j$ 
\State Construct effect signatures over $\mathcal{D}_{\mathrm{ft}}$ for the PCA-band neural sites
\For{$(\varepsilon,\beta_{\mathrm{neural}})\in\{0.5\}\times\{0.1,0.3,1,3\}$}
\State Compute the Sinkhorn UOT coupling in $\R^{2\times b}$ using $\varepsilon$ and $\beta_{\mathrm{neural}}$
\For{$Z\in\{\mathrm{AP},\mathrm{AT}\}$ with top layer $\ell$}
\For{$(\text{top-}K,\lambda)\in\{1,2,3,4,5\}\times\{0.5,1,2,4,8,12,20,30\}$}
\If{handle size does not exceed 1152 PCA directions}
\State Evaluate the candidate intervention handle $H_Z^{K,\lambda}$ on $\mathcal{D}_{\mathrm{cal}}$
\EndIf
\EndFor
\EndFor
\EndFor
\EndFor
\EndFor
\State Evaluate the best calibrated handles $H_{\mathrm{AP}}^\star,H_{\mathrm{AT}}^\star$ on $\mathcal{D}_{\mathrm{te}}$
\State End timer and report \twoPLOTpca{} runtime
\end{algorithmic}
\end{algorithm}

\begin{algorithm}[htbp]
\caption{MCQA \PLOTdas{}}
\label{alg:mcqa-plot-das}
\begin{algorithmic}[1]
\Require DAS learning rate $\mathrm{lr}=10^{-3}$, DAS maximum epochs $N=100$
\Require Splits $\mathcal{D}_{\mathrm{ft}}$, $\mathcal{D}_{\mathrm{cal}}$, and $\mathcal{D}_{\mathrm{te}}$, factual pre-trained Gemma-2-2B
\State Start timer
\State \underline{\textbf{Stage 1:}}
\State Construct effect signatures over $\mathcal{D}_{\mathrm{ft}}$ for final-token residual layers $\ell_0,\ldots,\ell_{25}$
\State Construct effect signatures over $\mathcal{D}_{\mathrm{ft}}$ for abstract variables $\mathrm{AP},\mathrm{AT}$
\For{$(\varepsilon,\beta_{\mathrm{neural}})\in\{0.5,1,2,4\}\times\{0.1,0.3,1,3\}$}
\State Compute the Sinkhorn UOT coupling in $\R^{2\times 26}$ using $\varepsilon$ and $\beta_{\mathrm{neural}}$
\For{$Z\in\{\mathrm{AP},\mathrm{AT}\}$}
\State Extract the six highest-mass layers for row $Z$
\State Evaluate their full-layer interventions for $Z$ on $\mathcal{D}_{\mathrm{cal}}$; retain highest-scoring layer $\ell_Z$
\EndFor
\EndFor
\State Choose $\{\ell_\mathrm{AP}, \ell_\mathrm{AT}\}$ that achieves highest mean calibration score across choices of $\varepsilon$ and $\beta_\mathrm{neural}$
\State \underline{\textbf{Stage 2:}}
\For{$Z \in \{\mathrm{AP},\mathrm{AT}\}$}
    \For{$k \in \{16,32,64,96,128,256,512,768,1024,1152\}$}
        \State Train a DAS rotation for $Z$ on $\mathcal{D}_{\mathrm{ft}}$ at layer $\ell_Z$ with subspace intervention size $k$
        \State Evaluate the rotation handle for $Z$ on $\mathcal{D}_{\mathrm{cal}}$
    \EndFor
\EndFor
\State Evaluate the best calibrated rotations for $\mathrm{AP}$ and $\mathrm{AT}$ on $\mathcal{D}_{\mathrm{te}}$
\State End timer and report \PLOTdas{} runtime
\end{algorithmic}
\end{algorithm}

\begin{algorithm}[htbp]
\caption{MCQA Full DAS}
\label{alg:mcqa-full-das}
\begin{algorithmic}[1]
\Require DAS learning rate $\mathrm{lr}=10^{-3}$, DAS maximum epochs $N=100$
\Require Splits $\mathcal{D}_{\mathrm{ft}}$, $\mathcal{D}_{\mathrm{cal}}$, and $\mathcal{D}_{\mathrm{te}}$, factual pre-trained Gemma-2-2B
\State Start timer
\For{$Z \in \{\mathrm{AP},\mathrm{AT}\}$}
    \For{$\ell \in \{\ell_0,\ldots,\ell_{25}\}$}
        \For{$k \in \{16,32,64,96,128,256,512,768,1024,1152\}$}
            \State Train a DAS rotation on $\mathcal{D}_{\mathrm{ft}}$ at layer $\ell$ with subspace intervention size $k$
            \State Evaluate the rotation handle for $Z$ on $\mathcal{D}_{\mathrm{cal}}$
        \EndFor
    \EndFor
\EndFor 
\State Evaluate the best calibrated rotation for $\mathrm{AP}$ and $\mathrm{AT}$ on $\mathcal{D}_{\mathrm{te}}$
\State End timer and report Full DAS runtime
\end{algorithmic}
\end{algorithm}

\newpage

\section{Additional Ablations}
\label{app:ablations}
\subsection{Non-OT Matching Alternatives}
\label{app:ablations-matching}

Although we use OT as the matching method of choice throughout the paper, this section compares it with two non-OT alternatives for defining the matching weights $\Pi$:
\begin{itemize}[leftmargin=1.3em,itemsep=0.7pt]
\vspace{-2mm}

\item \textbf{Cosine matching:}
Collect effect signatures as before, then define pairwise scores using their cosine similarity:
\[
A_{ij}^{\mathrm{cos}}
:=
\frac{u_i^\intercal v_j}
{\|u_i\|_2\|v_j\|_2},
\]
for nonzero signatures $u_i$ and $v_j$.

\item \textbf{IIA matching:}
For each abstract variable $i$ and neural site $j$, directly compute their alignment quality using interchange intervention accuracy on $\mathcal{D}_{\mathrm{ft}}$:
\[
A_{ij}^{\mathrm{IIA}}
:=
\frac{1}{T_{\mathrm{ft}}}
\sum_{t=1}^{T_{\mathrm{ft}}}
\mathds{1}_{\left\{
y_{j,t}^{\mathrm{nn\_swap}}
\sim
y_{i,t}^{\mathrm{abs\_swap}}
\right\}}.
\]

\end{itemize}

For either score matrix $A$, we apply a row-wise softmax to obtain the matching weights used for localization and handle calibration:
\[
\Pi_{ij}
=
\frac{\exp(A_{ij})}
{\sum_{j'=1}^{n}\exp(A_{ij'})}.
\]

We compare UOT, cosine, and IIA matching for \twoPLOTnat{} on MCQA, applying each method during both Stage A layer localization and Stage B within-layer native handle extraction. For UOT and cosine, we evaluate both the default family-wise signature used in the main text and the KL signature, whose construction is given in \Cref{app:ablations-signatures} below. IIA matching instead uses fit-bank intervention accuracy directly and does not require effect signatures. Because IIA is also our evaluation metric, this construction is intentionally evaluation-aligned, while test IIA remains held out. We therefore view it as a strong non-OT comparator.

\Cref{tab:matching-method-ablation} displays the results over five seeds. KL-based UOT achieves the highest mean overall IIA at $0.8620$. UOT also outperforms cosine under both signatures, yielding $0.8330$ versus $0.8000$ with the family-wise signature and $0.8620$ versus $0.8305$ with KL. IIA matching reaches mean IIA $0.8530$, between family-wise UOT and KL-based UOT. Cosine matching is faster than UOT under both signatures and avoids its transport-parameter sweeps (namely, $\varepsilon$ and $\beta$), while IIA matching lies between them in runtime. Thus, non-OT alternatives also produce accurate handles within the same progressive, gradient-free pipeline, with different accuracy--runtime tradeoffs. We adopt OT as the default because it provides a global constrained matching formulation, accommodates marginal relaxation through UOT, and draws on well-developed statistical and computational foundations, while the ablation shows competitive empirical performance.


\begin{table}[htbp]
\centering
\scriptsize
\setlength{\tabcolsep}{1.8pt}
\renewcommand{\arraystretch}{1.06}
\caption{\twoPLOTnat{} MCQA matching ablation over five seeds. Values are mean $\pm$ standard deviation. Runtime is serial time; layer/dim columns report the selected seed-0 handles.}
\vspace{-3mm}
\label{tab:matching-method-ablation}

\resizebox{\linewidth}{!}{%
\begin{tabular}{lcccccc}
\toprule
Method & AP & AT & Avg. & Runtime (min) & AP layer/dim & AT layer/dim \\
\midrule

UOT (family-wise)
& $0.7630 \pm 0.1026$
& $0.9030 \pm 0.0231$
& $0.8330 \pm 0.0523$
& $49.11 \pm 3.93$
& L17/512
& L24/1024 \\

Cosine (family-wise)
& $0.6960 \pm 0.0820$
& $0.9040 \pm 0.0258$
& $0.8000 \pm 0.0359$
& $\mathbf{15.60 \pm 0.42}$
& L19/768
& L24/1152 \\

\midrule

UOT (KL)
& $\mathbf{0.8250 \pm 0.0298}$
& $0.8990 \pm 0.0216$
& $\mathbf{0.8620 \pm 0.0174}$
& $40.85 \pm 1.92$
& L18/512
& L24/1152 \\

Cosine (KL)
& $0.7580 \pm 0.0635$
& $0.9030 \pm 0.0275$
& $0.8305 \pm 0.0349$
& $16.50 \pm 0.60$
& L18/1152
& L24/1152 \\

\midrule

IIA matching
& $0.7850 \pm 0.0491$
& $\mathbf{0.9210 \pm 0.0279}$
& $0.8530 \pm 0.0161$
& $23.79 \pm 0.72$
& L18/768
& L24/1024 \\

\bottomrule
\end{tabular}%
}
\vspace{-3mm}
\end{table}

\subsection{Effect Signature Ablation}
\label{app:ablations-signatures}

As an ablation study, we compare three effect signatures on MCQA: the family-wise signature on a truncated alphabet vocabulary used in the main text, a total-variation (TV) signature on the full output vocabulary, and a Kullback--Leibler (KL) signature on the full output vocabulary.

\begin{itemize}[leftmargin=1.3em,itemsep=0.7pt]
\vspace{-2mm}

\item \textbf{Family-wise signature (default):}
First, we define the abstract effect on fit pair $t$ as
\[
\Delta_{i,t}^{\mathrm{abs}}
=
y_{i,t}^{\mathrm{abs\_swap}}
-
y_t^{\mathrm{abs}}
\in \R^{26},
\]
where each abstract output is a one-hot encoding over the uppercase letters A--Z. We define the effect of neural site $s_j$ using raw logits:
\[
\Delta_{j,t}^{\mathrm{nn}}
=
y_{j,t}^{\mathrm{nn\_swap}}
-
y_t^{\mathrm{nn}}
\in \R^{26},
\]
where outputs are Gemma-2-2B's logits restricted to the tokens of the uppercase letters A--Z.

The $\operatorname{Agg}$ operation is defined as follows. Let $I_f$ contain the fit pairs from $\mathcal{D}_{\mathrm{ft}}$ in counterfactual family $f$, defined in \Cref{appen:mcqa}. For both abstract and neural effects, we average within each family, center across the $26$ coordinates, and $\ell_2$-normalize:
\[
\bar\Delta_f
=
\frac{1}{|I_f|}
\sum_{t\in I_f}\Delta_t,
\qquad
\widetilde\Delta_f
=
\bar\Delta_f
-
\frac{\mathbf{1}^{\top}\bar\Delta_f}{26}\mathbf{1},\qquad
z_f
=
\begin{cases}
\widetilde\Delta_f/\|\widetilde\Delta_f\|_2,
& \|\widetilde\Delta_f\|_2>0,\\
\widetilde\Delta_f,
& \text{otherwise}.
\end{cases}
\]
Finally, we concatenate the three family blocks, giving
\[
u_i^{\mathrm{fam}}
=
\left[
z_{i,\mathrm{answerPosition}}^{\mathrm{abs}},
z_{i,\mathrm{randomLetter}}^{\mathrm{abs}},
z_{i,\mathrm{answerPosition\_randomLetter}}^{\mathrm{abs}}
\right]
\in \R^{78},
\]
and
\[
v_j^{\mathrm{fam}}
=
\left[
z_{j,\mathrm{answerPosition}}^{\mathrm{nn}},
z_{j,\mathrm{randomLetter}}^{\mathrm{nn}},
z_{j,\mathrm{answerPosition\_randomLetter}}^{\mathrm{nn}}
\right]
\in \R^{78}.
\]

\medskip
\item \textbf{TV signature:}
We define the abstract effect signature to record whether the abstract intervention changes the output:
\[
u_i^{\mathrm{TV}}
=
\left[
\mathds{1}_{\left\{
y_{i,t}^{\mathrm{abs\_swap}}
\neq
y_t^{\mathrm{abs}}
\right\}}
\right]_{t=1}^{T_{\mathrm{ft}}}
\in \R^{200},
\]
where $T_{\mathrm{ft}}=200$. The intervention effect of neural site $s_j$ on fit pair $t$ is
\[
\Delta_{j,t}^{\mathrm{TV}}
=
\frac{1}{2}
\left\|
\operatorname{softmax}\!\left(y_{j,t}^{\mathrm{nn\_swap}}\right)
-
\operatorname{softmax}\!\left(y_t^{\mathrm{nn}}\right)
\right\|_1,
\]
where the neural logits and softmax distributions now span the full output vocabulary. Letting $\operatorname{Agg}$ concatenate these effects in fit-pair order gives the neural signature
\[
v_j^{\mathrm{TV}}
=
\left[
\Delta_{j,1}^{\mathrm{TV}},
\Delta_{j,2}^{\mathrm{TV}},
\ldots,
\Delta_{j,T_{\mathrm{ft}}}^{\mathrm{TV}}
\right]
\in \R^{200}.
\]

\medskip
\item \textbf{KL signature:}
The abstract signature is the same output-change indicator vector,
\[
u_i^{\mathrm{KL}}
=
u_i^{\mathrm{TV}}
\in \R^{200}.
\]
We define the effect of neural site $s_j$ on fit pair $t$ as
\[
\Delta_{j,t}^{\mathrm{KL}}
=
1-\exp\!\left(
-
\mathsf{KL}\!\left(
\operatorname{softmax}\!\left(y_{j,t}^{\mathrm{nn\_swap}}\right)
\,\middle\|\,
\operatorname{softmax}\!\left(y_t^{\mathrm{nn}}\right)
\right)
\right),
\]
with softmax distributions over the full output vocabulary. We use the transformation $1-\exp(-D_{\mathrm{KL}})$ to bound the raw KL value to $[0,1)$, then concatenate via $\operatorname{Agg}$ to obtain
\[
v_j^{\mathrm{KL}}
=
\left[
\Delta_{j,1}^{\mathrm{KL}},
\Delta_{j,2}^{\mathrm{KL}},
\ldots,
\Delta_{j,T_{\mathrm{ft}}}^{\mathrm{KL}}
\right]
\in \R^{200}.
\]

\end{itemize}

These variants differ in their output vocabulary, effect statistic, and aggregation, so the ablation compares complete signature constructions.

We compare the performance of these signatures under one-stage and two-stage PLOT in native and PCA representations, with TV reported only for two-stage native PLOT. For two-stage methods, each signature is used both for coarse layer localization and within-layer handle extraction. \Cref{tab:mcqa-signature-ablation} displays the results over five seeds.

With two-stage native PLOT, KL attains the highest mean test IIA at $0.8620$, compared with $0.8440$ for TV and $0.8330$ for the family-wise signature. With two-stage PCA, KL gives a smaller increase, from $0.8610$ to $0.8655$. In both representations, KL yields lower observed cross-seed variability in overall IIA and lower mean runtime than the family-wise signature.

Without staging, however, KL reduces mean IIA from $0.5955$ to $0.0085$ in native coordinates and from $0.8380$ to $0.4840$ in PCA. Thus, all three tested signatures yield accurate handles with two-stage PLOT, but signature choice interacts strongly with localization strategy.

\begin{table}[htbp]
\centering
\scriptsize
\setlength{\tabcolsep}{3pt}
\renewcommand{\arraystretch}{1.06}
\caption{MCQA effect-signature ablation with UOT over five seeds.
Values are mean $\pm$ standard deviation.
Family-wise is the default signature used in the main text;
TV is reported only for two-stage native PLOT.
Bold marks the highest mean overall IIA within each method group.}
\vspace{-3mm}
\label{tab:mcqa-signature-ablation}

\resizebox{\linewidth}{!}{%
\begin{tabular}{llcccc}
\toprule
Method & Signature & AP & AT & Avg. & Runtime (min) \\
\midrule

\onePLOTnat{}
& Family-wise
& $0.4290 \pm 0.3381$
& $0.7620 \pm 0.2958$
& $\mathbf{0.5955 \pm 0.2779}$
& $112.85 \pm 2.64$ \\

& KL
& $0.0100 \pm 0.0035$
& $0.0070 \pm 0.0057$
& $0.0085 \pm 0.0034$
& $131.06 \pm 4.07$ \\

\midrule

\onePLOTpca{}
& Family-wise
& $0.7640 \pm 0.1130$
& $0.9120 \pm 0.0561$
& $\mathbf{0.8380 \pm 0.0435}$
& $124.71 \pm 3.09$ \\

& KL
& $0.5970 \pm 0.2690$
& $0.3710 \pm 0.1104$
& $0.4840 \pm 0.1559$
& $140.87 \pm 4.03$ \\

\midrule

\twoPLOTnat{}
& Family-wise
& $0.7630 \pm 0.1026$
& $0.9030 \pm 0.0231$
& $0.8330 \pm 0.0523$
& $49.11 \pm 3.93$ \\

& TV
& $0.7820 \pm 0.0609$
& $0.9060 \pm 0.0282$
& $0.8440 \pm 0.0423$
& $41.04 \pm 1.60$ \\

& KL
& $0.8250 \pm 0.0298$
& $0.8990 \pm 0.0216$
& $\mathbf{0.8620 \pm 0.0174}$
& $40.85 \pm 1.92$ \\

\midrule

\twoPLOTpca{}
& Family-wise
& $0.7640 \pm 0.1099$
& $0.9580 \pm 0.0115$
& $0.8610 \pm 0.0526$
& $67.50 \pm 4.34$ \\

& KL
& $0.7710 \pm 0.0323$
& $0.9600 \pm 0.0117$
& $\mathbf{0.8655 \pm 0.0152}$
& $61.73 \pm 1.63$ \\

\bottomrule
\end{tabular}%
}
\vspace{-3mm}
\end{table}

\end{document}